\documentclass[letterpaper]{article}
\usepackage[preprint]{aaai2027}
\usepackage[hyphens]{url}
\usepackage{graphicx}
\usepackage{amsmath,amssymb}
\usepackage{natbib}
\usepackage{caption}
\usepackage{booktabs}
\usepackage{array}
\usepackage{longtable}
\usepackage{tabularx}
\usepackage{microtype}
\usepackage{enumitem}
\setlist{nosep,leftmargin=1.4em}

\title{Diagnosing Correctness Probes under Self-Judgement Confounding}
\author{
    Yi-Long Lu\textsuperscript{\rm 1,\rm 2}
    }
\affiliations{
    \textsuperscript{\rm 1}Key Laboratory of Adolescent Cyberpsychology and Behavior (CCNU),\\
    Ministry of Education, Wuhan 430079, China;\\
    \textsuperscript{\rm 2}Key Laboratory of Human Development and Mental Health of Hubei Province, School of Psychology,\\
    Central China Normal University, Wuhan 430079, China;\\
    luyilong@ccnu.edu.cn
    }

\begin{document}

\maketitle
\begin{abstract}
Hidden-state readouts can predict whether language-model outputs are correct,
but objective correctness (OC) usually agrees with the model's own
self-judgement (SJ), leaving the decoded signal semantically ambiguous. We
construct conflict cases in which OC and SJ predict opposite readout orderings.
On high-confidence
disagreements, conventional correctness-labelled contrasts often rank
incorrect/self-endorsed responses above correct/self-rejected responses,
following SJ rather than OC. We estimate factorial SJ- and OC-associated
directions and evaluate their polarity across mathematical reasoning and
factual recall. Across four instruction-tuned models up to 14B parameters, the
SJ-associated direction transfers above chance in both cross-domain directions for every model,
whereas the OC-associated direction has a below-chance point estimate for the
expected OC ordering in every corresponding condition. This transfer asymmetry
develops across middle-to-late layers,
persists under answer-likelihood, sequence-length, and
null-direction controls, and extends to MMLU and binary TruthfulQA without
target-domain direction fitting. Across the studied models and diagnostic
subsets, the most reliably transferable component preserves SJ-associated
polarity. Transferability alone therefore does not establish
objective-correctness semantics.
\end{abstract}

\section{Introduction}

Large language models (LLMs) can produce fluent answers that are factually
wrong, motivating efforts to read factual reliability directly from their
internal representations \cite{lin2022truthfulqa,huang2025surveyHallucinationLLMs}.
Early work showed that latent knowledge and statement truthfulness are often
decodable from hidden activations
\cite{burns2023discovering,azaria2023internal,li2023inference}, while subsequent
analyses reported a simple linear geometry for factual truth across datasets
\cite{marks2024geometry}. Some of these readouts transfer across logical
transformations, question-answering tasks, and external knowledge settings
\cite{bao2025geometryTruth}, suggesting that truth-related information may be
encoded in a reusable form. Yet reusability alone does not determine which
variable a readout represents.\footnote{Code and processed data are available at
\url{https://github.com/Yilong-Lu/Diagnosing_Correctness}.}

The central issue is therefore not only whether a truth-related direction
transfers, but which variable preserves its polarity under transfer. Such
directions vary with layer, task type, task complexity, and instructions
\cite{azizian2025orthogonalTruthGeometries,poulis2026limitsTruthDirections};
recent evidence places broadly shared and domain-specific directions on a
continuum \cite{ying2026truthfulnessSpectrum}. Separability can degrade under
distribution shift, and some factual errors share internal geometry with
successful knowledge recall
\cite{haller2025knowledgeBrittle,cheang2026doLLMsReallyKnow}. These findings
leave the semantic interpretation of a transferable correctness readout unresolved.

\begin{figure*}[t]
\centering
\includegraphics[width=0.99\textwidth]{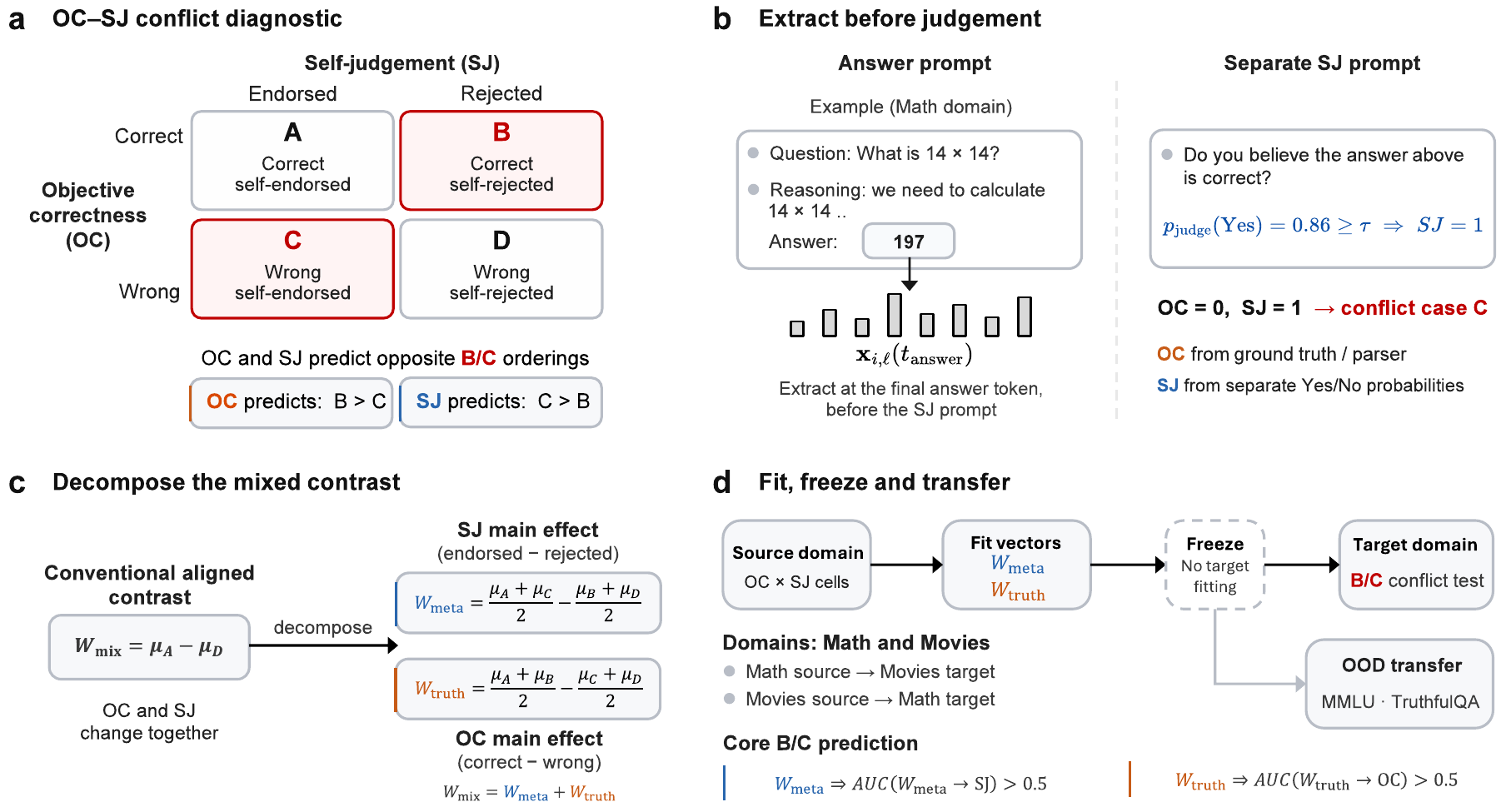}
\caption{Conflict-based semantic validation of correctness probes.
(a) Objective correctness (OC) and self-judgement (SJ) define four response
types. On B/C conflicts, OC predicts $B>C$, whereas SJ predicts $C>B$.
(b) Hidden states are extracted at the final answer token, before the separate
judgement prompt used to define SJ.
(c) The conventional contrast
$W_{\mathrm{mix}}=\mu_A-\mu_D$ decomposes into the factorial SJ- and
OC-associated directions $W_{\mathrm{meta}}$ and $W_{\mathrm{truth}}$.
(d) Source-domain directions are frozen and evaluated on B/C conflicts in
another domain, including zero-target-fitting transfer to MMLU and binary
TruthfulQA. Successful component transfer corresponds to
$\mathrm{AUC}(W_{\mathrm{meta}}\!\rightarrow\!\mathrm{SJ})>0.5$ and
$\mathrm{AUC}(W_{\mathrm{truth}}\!\rightarrow\!\mathrm{OC})>0.5$.}
\label{fig:concept}
\end{figure*}

One candidate source of this ambiguity is the model's own evaluative stance.
Recent work reports a shared representational dimension spanning subjective
evaluation and assent to factual claims
\cite{lu2025unifiedRepresentationJudgmentLLMs}. A broader literature studies
response evaluation through behavioural confidence, failure prediction, and
internal activation analyses
\cite{kadavath2022language,wang2025decouplingMetacognitionCognition,
jian2025metacognitiveActivations,kumaran2026verbalConfidence}; aggregating
self-evaluation-related states can also improve uncertainty calibration
\cite{xiao2026eagle}. At the feature level, correctness and output uncertainty
can depend on partly distinct representations
\cite{patel2026functionalDissociation}. These findings motivate a sharper
diagnostic question for truth probing: when a correctness-labelled readout
transfers, does it preserve the relation between an answer and external ground
truth, or the model's own evaluation of that answer?

To make this distinction operational, we separate two response-level
variables. Objective correctness (OC) records whether an answer is externally
scored as correct. Self-judgement (SJ) records whether the model subsequently
judges that answer to be correct; this correctness-directed, second-order
assessment serves as our operational measure of metacognitive judgement
\cite{steyvers2026metacognitionUncertaintyCommunication}. Because OC and SJ
usually agree, a conventional correct-versus-incorrect contrast can conflate
their activation correlates. Their competing interpretations become testable
when the two variables disagree.

The empirical design crosses OC with SJ to form four response types. The
critical comparison is between objectively correct answers that the model
rejects (B) and objectively wrong answers that it endorses (C). The two
candidate signals predict opposite orderings: an OC-associated direction should
rank B above C, whereas an SJ-associated direction should rank C above B.
Even an OC-only mass-mean control fitted without SJ labels follows SJ in all
eight cross-domain comparisons. Factorial OC- and SJ-associated directions then
assess which polarity transfers.
Across four instruction-tuned LLMs, the SJ-associated $W_{\mathrm{meta}}$
direction has the expected polarity in every cross-domain
condition spanning mathematical reasoning, factual recall, MMLU, and binary
TruthfulQA, whereas the OC-associated
$W_{\mathrm{truth}}$ direction does not. Self-judgement is therefore a
substantial source of semantic ambiguity in transferable correctness
readouts.

\section{Related Work}

\paragraph{Truth-direction transfer and semantic validation.}
Latent knowledge and factual truthfulness are often decodable from LLM
activations using unsupervised, linear, or nonlinear readouts
\cite{burns2023discovering,azaria2023internal,li2023inference,
marks2024geometry}. Some truth directions generalize across logical
transformations and QA settings \cite{bao2025geometryTruth}; others depend
strongly on layer, task family, instructions, or truth type
\cite{azizian2025orthogonalTruthGeometries,poulis2026limitsTruthDirections,
ying2026truthfulnessSpectrum,schouten2025contextSensitiveTruth}. Apparent truth
signals can also reflect superficial task features or knowledge recall
\cite{haller2025knowledgeBrittle,cheang2026doLLMsReallyKnow}. Their
generalization to realistic model-generated responses remains challenging
\cite{servedio2025hiddenStates}. Query--probe disagreement can reflect
calibration or heterogeneous errors \cite{liu2023cognitiveDissonance}, while
inter-model disagreement can reveal domain-specific privileged correctness
information \cite{ashuach2026masked}. Predictive decodability therefore does
not identify the variable a readout exploits
\cite{hewitt2019designing,belinkov2021probing}; instructions and interaction can
further redirect truth-related representations
\cite{long2025truthfulRepresentationsFlip,wang2026truthOverriddenSycophancy}.
OC--SJ conflicts make semantic polarity directly testable because the two
variables prescribe opposite rankings.

\paragraph{Self-evaluation, uncertainty, and introspection.}
LLM self-evaluation has been studied through P(True), verbal confidence,
uncertainty elicitation, unknown detection, and failure prediction
\cite{kadavath2022language,lin2022teaching,tian2023just,xiong2023can,
yin2023knowDontKnow,wang2025decouplingMetacognitionCognition}. Hidden-state
trajectories have also been used to predict response correctness without an
explicit verbal report \cite{wang2025chainEmbedding}. Correctness and output
uncertainty can be functionally dissociated at the sparse-feature level
\cite{patel2026functionalDissociation}, while confidence-related information
around answer completion can exceed token log-probability
\cite{kumaran2026verbalConfidence,kumaran2026detectCorrectErrors}. Stronger
claims of privileged self-access face different tests: prompted reports can
fail to recover a model's own linguistic knowledge, and apparent introspective
performance can be reproduced from surface cues
\cite{song2025failIntrospect,singh2026introspectionRealityCheck}. Existing
studies have separately decoded correctness-related and response-evaluation
signals; we ask which correlated variable preserves its semantic polarity when
they conflict. SJ serves here as a correctness-directed behavioural label whose
answer-token correlate is evaluated on those conflict cases.

\section{Method}

\paragraph{Setup and labels.}
For an answered item $i$ at layer $\ell$, let $x_{i,\ell}$ be the hidden state
at the answer-specific final token. OC indicates objective correctness, and SJ
indicates the model's binary judgement of whether its own answer is correct.
The judgement turn asks, ``Do you believe the answer above is correct? Answer
only with Yes or No.'' Next-token Yes/No scores define the binary correctness
probability $p_{\mathrm{judge}}$; this separate turn is excluded from activation
extraction. The factorial contrasts below quantify the marginal activation
associations of OC and SJ across the observed response types.
Confidence is used only to exclude weak judgements: the main analyses retain
$p_{\mathrm{judge}}\leq0.3$ or $p_{\mathrm{judge}}\geq0.7$ and define SJ by the
retained side. This defines $A=(1,1)$, $B=(1,0)$, $C=(0,1)$, and $D=(0,0)$ over
(OC,SJ); $B\cup C$ is the conflict set where correctness and self-judgement disagree.

\paragraph{Factorial directions.}
Let $\mu_A,\mu_B,\mu_C,\mu_D$ be quadrant mean activations after source/train-only
centering. We compare an aligned mixed contrast with two factorial main effects:
\[
\begin{array}{rcl}
W_{\mathrm{mix}}&=&\mu_A-\mu_D,\\
W_{\mathrm{meta}}&=&[(\mu_A-\mu_B)+(\mu_C-\mu_D)]/2,\\
W_{\mathrm{truth}}&=&[(\mu_A-\mu_C)+(\mu_B-\mu_D)]/2,
\end{array}
\]
where $W_{\mathrm{meta}}$ and $W_{\mathrm{truth}}$ are operational labels for
the SJ-associated and OC-associated factorial contrasts, respectively.
The interpretation follows from a minimal associative representation model with signed labels
$o_i=2\mathrm{OC}_i-1$ and $s_i=2\mathrm{SJ}_i-1$:
\[
x_i=\alpha_t o_i v^{\mathrm{OC}}_t+\beta_t s_i v^{\mathrm{SJ}}_t
+\eta_t o_i s_i v^{\mathrm{INT}}_t+\epsilon_i .
\]
Here $v^{\mathrm{OC}}_t$ and $v^{\mathrm{SJ}}_t$ are task-specific latent
directions associated with correctness and self-judgement, $v^{\mathrm{INT}}_t$
captures their interaction, and $\alpha_t,\beta_t,\eta_t$ are task-dependent
effect magnitudes.
Under this model, substituting the four cell means gives
$W_{\mathrm{meta}}=2\beta_t v^{\mathrm{SJ}}_t$,
$W_{\mathrm{truth}}=2\alpha_t v^{\mathrm{OC}}_t$, and
$W_{\mathrm{mix}}=W_{\mathrm{truth}}+W_{\mathrm{meta}}$; the interaction term
cancels in the main effects. On B/C conflicts, OC predicts $B>C$ whereas SJ
predicts $C>B$. The corresponding mean-score difference under the mixed
direction is
\[
\begin{array}{rcl}
\Delta_{\mathrm{mix}}&=&
\mathrm{E}[\langle x,W_{\mathrm{mix}}\rangle|C]-
\mathrm{E}[\langle x,W_{\mathrm{mix}}\rangle|B]\\
&=&4\beta_t^2\|v^{\mathrm{SJ}}_t\|^2-
4\alpha_t^2\|v^{\mathrm{OC}}_t\|^2 .
\end{array}
\]
Thus, when these responses are scored by $W_{\mathrm{mix}}$, an AUC above 0.5
for ranking C over B indicates that the SJ-associated component dominates the
conflict-set ordering.
Cross-domain transfer then tests whether a source-estimated SJ-associated
direction preserves SJ polarity in the target domain more reliably than the
corresponding OC-associated direction preserves OC polarity.

\paragraph{Strict pairs.}
For each model--domain combination, we sample eight stochastic free-response
answers per question and retain questions with at least one correct and one
incorrect usable response. We randomly select one of each and ask the same
model to judge whether its response is correct. Under the strict threshold
$\tau=0.7$, a pair is retained only when both self-judgements satisfy the
symmetric high-confidence criterion. These analyses therefore characterize a
high-confidence diagnostic subset of questions exhibiting response variability;
they do not estimate the prevalence of B/C cases in unconstrained model outputs.
Across model--domain cells, the resulting sets contain 362--2,775 retained
questions and 249--2,146 B/C conflict responses.
Full attrition statistics and quadrant counts are reported in the supplement.

\paragraph{Activation site.}
Activations are extracted from the answer prompt together with the
corresponding assistant answer: the final token of the numeric answer span for
Math, the final token of the actor-name answer for Movies, and the forced
answer-letter token for OOD.
At layer $\ell$, $x_{i,\ell}$ denotes the hidden state at this token after transformer block $\ell$,
that is, the post-block residual-stream state. The judgement prompt is used only to compute
$p_{\mathrm{judge}}$; abbreviated prompt cores and full templates are provided
in the supplement.

\begin{figure*}[t]
\centering
\includegraphics[width=0.99\textwidth]{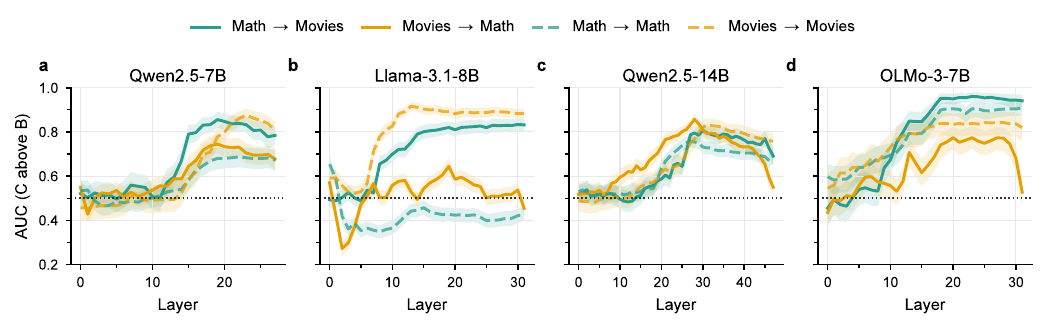}
\caption{Traditional mixed direction on OC--SJ conflicts. Curves show the AUC
for ranking C (wrong/self-endorsed) above B (correct/self-rejected) across
layers. Values above 0.5 indicate that
$W_{\mathrm{mix}}$ ranks wrong/self-endorsed samples above correct/self-rejected
samples. Directions are derived on Math (teal) or Movies (amber) datasets; solid and dashed lines denote
cross-domain and within-domain evaluation, respectively. Shaded bands denote bootstrap 95\% CIs conditional
on the fitted source direction; all conflict responses from a question are
resampled together.}
\label{fig:wmix}
\end{figure*}

\begin{figure*}[t]
\centering
\includegraphics[width=0.99\textwidth]{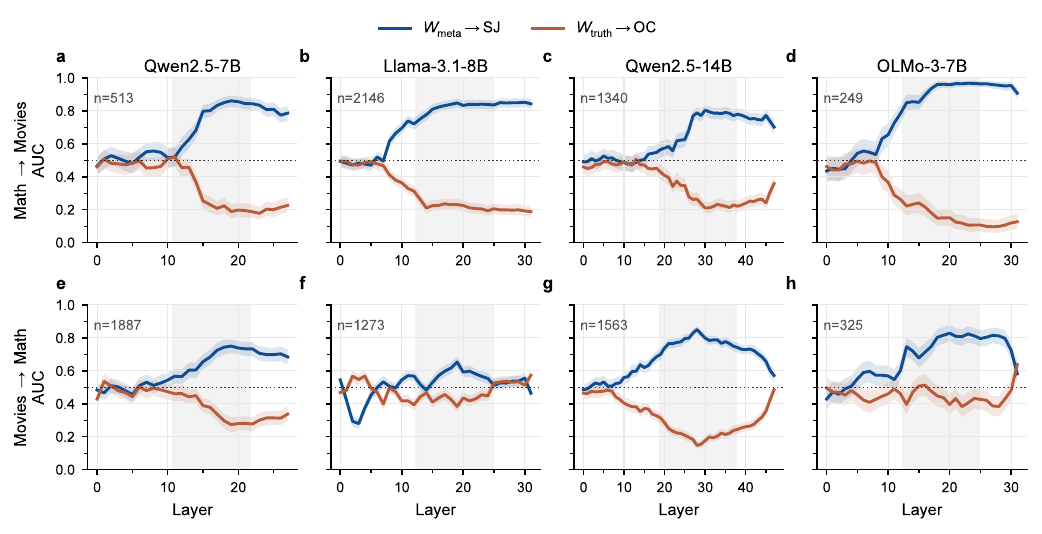}
\caption{Cross-domain component transfer across layers. Directions fitted in
one domain are evaluated on B/C conflicts in the other; rows show transfer
directions and columns show models. Blue curves show
$\mathrm{AUC}(W_{\mathrm{meta}}\!\rightarrow\!\mathrm{SJ})$, and brown-orange
curves show $\mathrm{AUC}(W_{\mathrm{truth}}\!\rightarrow\!\mathrm{OC})$. The dotted
horizontal line marks chance, and the pale vertical band marks the fixed
normalized-depth window $[0.40,0.80]$ summarized in
Table~\ref{tab:exp2b}. Shaded bands are 95\% target-question
cluster-bootstrap CIs conditional on the fitted source direction; in-panel
$n$ denotes the number of target B/C responses contributing to each
layer-wise AUC.}
\label{fig:exp2b}
\end{figure*}

\section{Experimental Protocol}

\paragraph{Datasets and models.}
For mathematical reasoning, we curate a Math dataset from an initial pool of
16,700 questions collected from public sources, including GSM8K and
MATH-style items \cite{cobbe2021training,hendrycks2021math}.
A Qwen2.5-1.5B-Instruct pilot removes questions solved on all eight
attempts and the longest approximately 10\% of responses, yielding 5,549
questions. This filtering preserves response variability while limiting
extreme generation lengths.
Movies is a person-name factual-recall domain using 17,856 actor-question
records from the Movies QA dataset described by \citet{orgad2024llmsKnowMore}.
OC is determined by domain-specific numeric-answer parsing for Math and normalized actor-name matching for Movies.
OOD evaluations use 4-choice MMLU \cite{hendrycks2021measuring} and binary TruthfulQA
\cite{lin2022truthfulqa}. MMLU contributes four options per question and binary
TruthfulQA two, with dataset answer keys defining OC. Candidate letters are
inserted as assistant responses and judged with the same SJ query; neither
target dataset contributes to direction fitting. We evaluate four instruction
models with at most 14B parameters: Qwen2.5-7B-Instruct, Llama-3.1-8B-Instruct,
Qwen2.5-14B-Instruct, and OLMo-3-7B-Instruct
\cite{qwen2024qwen25,ai2024llama3,teamolmo2025olmo3}.
Full details appear in the supplement.

\paragraph{Implementation.}
For each model, we performed forward passes in bfloat16 and extracted hidden
states at the final answer token.
All subsequent analyses were performed offline using the saved activations.
We set the temperature to 1.0 for both main-model pass@8 generation and the
one-token Yes/No judgement. Maximum generation lengths are 384 tokens for Math
and 256 for Movies. The resulting judgement score defines SJ.

\paragraph{Experimental comparisons.}
Evaluation proceeds from a conflict-set diagnostic to within-domain
validation and cross-domain transfer.
Experiment 1 tests whether the B/C ordering
induced by the conventional mixed direction $W_{\mathrm{mix}}$ follows OC or SJ
when the two variables disagree.
We compute the binary AUC with C (wrong/self-endorsed) as the positive class
and B (correct/self-rejected) as negative class. Values above 0.5 indicate
that C responses receive higher scores, opposite to the
ordering predicted by an OC-specific direction. The OC-only mass-mean control
forms a canonical unwhitened class-mean contrast from all retained source rows,
$W_{\mathrm{OC-only}}=\mathrm{E}[x\mid\mathrm{OC}=1]
-\mathrm{E}[x\mid\mathrm{OC}=0]$,
without using SJ labels \cite{marks2024geometry,zou2023representation}.
Because each retained question contributes one correct and one incorrect
response, this contrast is equivalent to the mean within-question
correct-minus-incorrect activation difference.

Experiment 2A tests whether the factorial directions preserve their expected
SJ and OC rankings on held-out B/C conflicts within domain. We use five
question-grouped folds, fitting centering and both directions on training questions
and scoring their held-out responses. Experiment 2B assesses bidirectional
cross-domain transfer by fitting the directions on all retained rows in one source
domain, freezing them, and evaluating them on B/C conflicts in the other domain:
$W_{\mathrm{meta}}\!\rightarrow\!\mathrm{SJ}$ with
$W_{\mathrm{truth}}\!\rightarrow\!\mathrm{OC}$.

\paragraph{Primary estimands and inference.}
We quantify component transfer using two AUCs. Because OC and SJ are
complementary labels on B/C conflicts, we define
$
\Delta_{\mathrm{CB}}=\mathrm{AUC}(W_{\mathrm{meta}}\rightarrow\mathrm{SJ})
-\mathrm{AUC}(W_{\mathrm{truth}}\rightarrow\mathrm{OC})
$
to summarize the relative strength of the two component transfers.
Each component AUC is evaluated separately against chance ($0.5$),
testing whether $W_{\mathrm{meta}}$ ranks SJ and $W_{\mathrm{truth}}$ ranks OC
on the same target responses.
All-layer curves show layer-wise structure, with shaded bands from a
target-question bootstrap that keeps all conflict rows from a question
together. All reported intervals are two-sided 95\% confidence intervals (CIs).
All-layer and OOD intervals condition on the fitted source direction. Primary
Exp2B fixed-window inference independently resamples source and target questions,
refits both factorial directions in every replicate, and averages layer-wise AUCs over an
architecture-normalized middle-to-late window, selecting layers with
$\ell/\ell_{\max}\in[0.40,0.80]$. This fixed window excludes architecture
endpoints and avoids selecting a model-specific peak layer.
Fixed-window bootstrap estimates provide inference, whereas the full profiles
characterize depth-dependent structure.

\paragraph{Question-level controls.}
The source-side sensitivity analysis takes the within-question
correct-minus-incorrect activation difference, eliminating additive question
intercepts before direction fitting. At each layer, multivariate least squares
jointly estimates OC, SJ, and interaction coefficient vectors from these paired
differences; the resulting $\widehat b_{\mathrm{SJ}}$ and
$\widehat b_{\mathrm{OC}}$ replace $W_{\mathrm{meta}}$ and
$W_{\mathrm{truth}}$ in the unchanged fixed-window B/C evaluation. Independent
source- and target-question bootstrap resampling refits the adjusted directions.

The complementary target-side analysis keeps the pooled source directions
fixed and uses every retained target row. Component scores are standardized
within layer and target domain, averaged over the fixed window, and regressed
on signed OC, SJ, and their interaction with target-question fixed effects.
Its cross-component contrast is
$\Delta_{\mathrm{FE}}=\beta^{(W_{\mathrm{meta}})}_{\mathrm{SJ}}
-\beta^{(W_{\mathrm{truth}})}_{\mathrm{OC}}$. Target-question bootstrap
replicates repeat score standardization and model fitting. This all-cell
analysis restores the aligned cells while absorbing target-question-specific
differences. Within-component coefficient differences additionally test whether $W_{\mathrm{meta}}$ loads more strongly on SJ than OC, and $W_{\mathrm{truth}}$ on OC than SJ.

\paragraph{Main-domain controls.}
Source-label-shuffle nulls test whether transfer depends on the fitted source
label geometry, while norm-matched random directions test whether arbitrary
activation directions yield the same pattern. B/C matching on rendered-sequence
token count tests prompt-plus-response length composition, and symmetric confidence-threshold sensitivity
tests dependence on judgement certainty. For Math and Movies,
we also score the generated assistant response tokens. At each
layer, projection scores are residualized against mean response log-probability
and response-token count before computing component AUCs; the fixed-window statistic
then averages those layer-wise AUCs, matching the primary estimand. Each
bootstrap replicate refits the layer-wise nuisance regressions.

\paragraph{OOD controls.}
Forced-choice analyses add answer-letter fixed effects to separate the
projection signal from option identity. A target-side elicitation control replaces Yes/No with reversed X/Y mappings
to test sensitivity to the judgement-to-label assignment.
Their correctness-oriented log odds are averaged before applying the same symmetric confidence threshold;
the source directions remain fixed.

\begin{table*}[thbp]
\centering

\begin{tabular}{llccc}
\toprule
Model & Direction & $W_{\mathrm{meta}}\!\rightarrow$SJ AUC & $W_{\mathrm{truth}}\!\rightarrow$OC AUC & $\Delta_{\mathrm{CB}}$ \\
\midrule
Qwen2.5-7B & Math$\rightarrow$Movies & 0.750 [0.715, 0.782] & 0.300 [0.263, 0.339] & 0.450 [0.386, 0.513] \\
Qwen2.5-7B & Movies$\rightarrow$Math & 0.670 [0.629, 0.708] & 0.368 [0.332, 0.414] & 0.302 [0.219, 0.371] \\
Llama-3.1-8B & Math$\rightarrow$Movies & 0.823 [0.796, 0.848] & 0.222 [0.191, 0.252] & 0.601 [0.545, 0.655] \\
Llama-3.1-8B & Movies$\rightarrow$Math & 0.572 [0.540, 0.603] & 0.434 [0.400, 0.471] & 0.138 [0.073, 0.198] \\
Qwen2.5-14B & Math$\rightarrow$Movies & 0.708 [0.679, 0.736] & 0.286 [0.259, 0.321] & 0.422 [0.365, 0.470] \\
Qwen2.5-14B & Movies$\rightarrow$Math & 0.781 [0.753, 0.802] & 0.215 [0.196, 0.246] & 0.566 [0.513, 0.602] \\
OLMo-3-7B & Math$\rightarrow$Movies & 0.928 [0.899, 0.950] & 0.167 [0.119, 0.243] & 0.761 [0.667, 0.818] \\
OLMo-3-7B & Movies$\rightarrow$Math & 0.776 [0.726, 0.821] & 0.441 [0.379, 0.511] & 0.334 [0.229, 0.430] \\
\bottomrule
\end{tabular}
\caption{Exp2B fixed-window component transfer and relative asymmetry.
Estimates are averaged over normalized layer depths $[0.40,0.80]$ using
1,000 independent bootstrap resamples of source and target questions.}
\label{tab:exp2b}
\end{table*}

\section{Results}

\paragraph{Exp1: correctness-labelled contrasts reverse the OC ordering on conflicts.}
If $W_{\mathrm{mix}}$ were a clean OC readout, it should prefer B
samples (correct/self-rejected) over C samples (wrong/self-endorsed), despite
the model's opposite judgement. Equivalently, the pairwise AUC that treats C as
the positive class, $\Pr[s(C)>s(B)]$, should fall below 0.5.
Figure~\ref{fig:wmix} shows the opposite pattern: after the early layers, the
curves are usually above 0.5, indicating that $W_{\mathrm{mix}}$ often assigns
higher scores to wrong/self-endorsed answers than to correct/self-rejected
answers.

The OC-only mass-mean control strengthens this diagnosis without using SJ labels
to fit the source direction. Its fixed-window AUC for ranking C above B exceeds
0.5 in all eight model-by-direction cross-domain evaluations, with 95\% CIs above
chance and estimates ranging from 0.567 to 0.874. Correctness-labelled mean
contrasts can therefore follow the model's judgement on cases where judgement
and correctness disagree, even when SJ is absent from probe fitting.

\paragraph{Exp2A: the SJ-associated contrast is stable within domain.}
Question-grouped folds keep the paired responses to one question in the same
partition. In held-out questions, $W_{\mathrm{meta}}$ predicts SJ above chance
in all eight model--domain conditions (AUC 0.649--0.915), with every 95\% CI
above 0.5. By contrast, $W_{\mathrm{truth}}$ predicts OC above chance only for
Llama-3.1-8B Math; its other seven point estimates are below 0.5
(0.219--0.483). Consequently, $\Delta_{\mathrm{CB}}$ is positive in seven of
eight conditions (0.173--0.696), with all seven 95\% CIs above zero. The sole
negative contrast is Llama-3.1-8B Math, where both components discriminate
above chance and $W_{\mathrm{truth}}$ is stronger
($-0.042$ [$-0.097$, 0.012]). Exp2A therefore supports the expected in-domain
SJ ordering but does not support a generally stable OC-associated readout;
complete layer-wise results are reported in the supplement.

\begin{figure*}[t]
\centering
\includegraphics[width=0.99\textwidth]{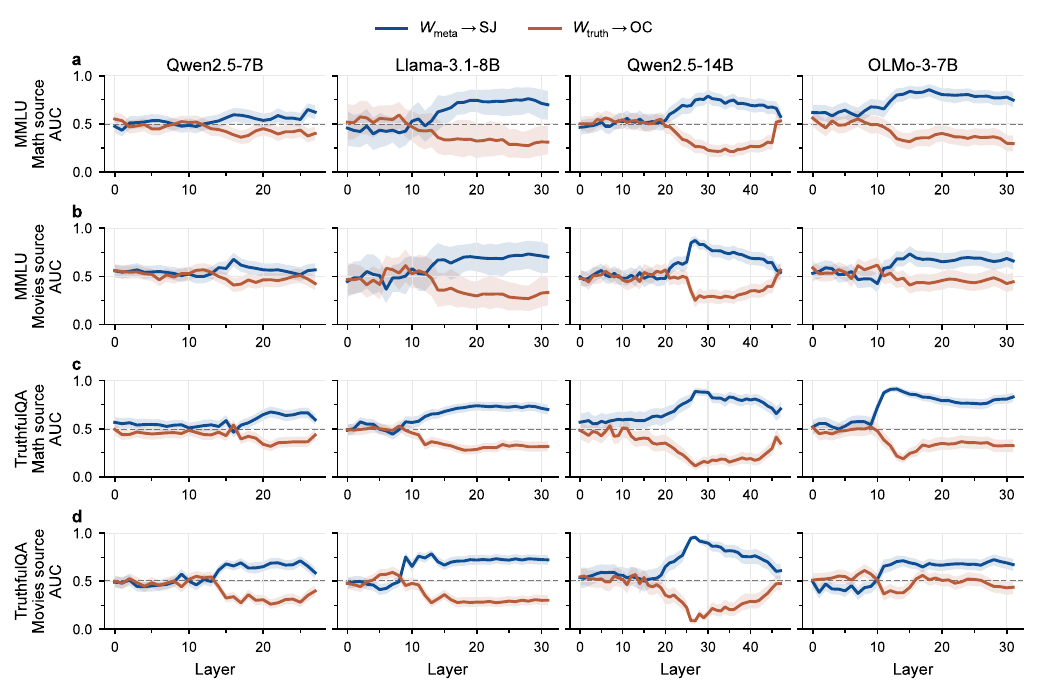}
\caption{OOD transfer without target-domain direction fitting. Rows show
(a) Math$\rightarrow$MMLU, (b) Movies$\rightarrow$MMLU,
(c) Math$\rightarrow$binary TruthfulQA, and
(d) Movies$\rightarrow$binary TruthfulQA; columns denote models.
Blue curves show
$\mathrm{AUC}(W_{\mathrm{meta}}\!\rightarrow\!\mathrm{SJ})$, and orange curves
show $\mathrm{AUC}(W_{\mathrm{truth}}\!\rightarrow\!\mathrm{OC})$ on OOD B/C
conflicts. The dashed line marks chance. Shaded bands show 95\% CIs from a target-question cluster bootstrap,
conditional on the fitted source direction.}
\label{fig:ood_components}
\end{figure*}

\paragraph{Exp2B: only the SJ-associated direction preserves cross-domain polarity.}
Directions fitted on Math are evaluated on Movies, and directions fitted on
Movies are evaluated on Math. The layer-wise profiles in
Figure~\ref{fig:exp2b} reveal a common progression. The component curves are
near chance in early layers and then separate across middle-to-late layers: the
SJ-associated curve rises above chance, whereas the OC-associated curve remains
near chance or reverses polarity. The separation is sustained through the
fixed analysis window rather than driven by a single isolated layer.
Across layers and transfer conditions, $\Delta_{\mathrm{CB}}>0$ in 255/280
(91\%) cross-domain model--transfer--layer rows.

The fixed-window estimates confirm this visual pattern. All eight
$W_{\mathrm{meta}}\rightarrow\mathrm{SJ}$ AUCs exceed 0.5 with joint bootstrap
95\% CIs above chance (0.572--0.928), whereas all eight
$W_{\mathrm{truth}}\rightarrow\mathrm{OC}$ point estimates are below chance
(0.167--0.441), with seven joint 95\% CIs excluding 0.5. The relative
$\Delta_{\mathrm{CB}}$ is positive for every model and transfer direction,
ranging from 0.138 to 0.761, with all eight joint source--target bootstrap 95\%
CIs above zero (Table~\ref{tab:exp2b}).

\paragraph{Question- and response-level controls.}
Source-side estimation from within-question activation differences preserves
the cross-domain separation: all eight adjusted $\Delta_{\mathrm{CB}}$ point
estimates are positive (0.130--0.748), with joint 95\% CIs excluding zero.
The complementary target-side all-cell analysis restores all four target cells
while absorbing target-question intercepts. Across all eight transfers,
$W_{\mathrm{meta}}$ has a positive SJ coefficient, and all
$\Delta_{\mathrm{FE}}$ estimates are positive (0.230--0.530), with 95\% CIs
excluding zero. Within both transferred scores, the SJ coefficient exceeds the
OC coefficient in every transfer, with all difference CIs excluding zero.
Thus, $W_{\mathrm{meta}}$ retains SJ specificity, whereas
$W_{\mathrm{truth}}$ does not retain OC specificity. Response-level controls
yield the same separation: residualizing projection scores against response
log-probability and response-token count leaves all eight fixed-window
$\Delta_{\mathrm{CB}}$ estimates positive (0.180--0.751), with CIs excluding
zero. Rendered-sequence B/C token-count matching yields a minimum estimate of
0.198 [0.065, 0.329].

\paragraph{Cross-domain transfer to MMLU and TruthfulQA.}
Directions estimated from Math and Movies are next evaluated on MMLU and binary
TruthfulQA without fitting on either target dataset. Each model--target
conflict set contains 198--548 candidate responses from 160--483 questions.
In the fixed window, all 16 $W_{\mathrm{meta}}\rightarrow\mathrm{SJ}$ AUCs
exceed 0.5 (0.544--0.823), with 15/16 95\% CIs above chance.
By contrast, all 16 $W_{\mathrm{truth}}\rightarrow\mathrm{OC}$ AUCs are below
0.5 (0.206--0.487), with 11 of the 16 corresponding 95\% CIs lying entirely
below chance.
The resulting $\Delta_{\mathrm{CB}}$ values range from 0.091 to 0.581.
Figure~\ref{fig:ood_components} shows the same component separation across
depth: the SJ-associated curves rise above chance in middle-to-late layers,
whereas the OC-associated curves remain near chance or reverse polarity.

\paragraph{Forced-choice and elicitation controls.}
The OOD separation is not explained by immediate preference for the forced
answer option. Residualization against answer-option log-probability and token
count leaves 15/16 estimates positive, with 15/16 95\% CIs above zero. Adding
answer-letter fixed effects yields the same counts. In both analyses, the sole
nonpositive estimate is
OLMo-3-7B Movies$\rightarrow$TruthfulQA, whose letter-controlled estimate is
$-0.030$ [$-0.138$, 0.077]. With counterbalanced X/Y judgement labels, all 16
fixed-window estimates remain positive (0.176--0.682), and
12/16 95\% CIs lie above zero. The four inconclusive CIs are the two source
directions into Llama-3.1-8B or OLMo-3-7B MMLU cells, where the balanced
conflict sets contain only 13 or 4 B responses, respectively.

\paragraph{Null and sensitivity analyses.}
The observed window means exceed the corresponding 95\% null intervals in all
16 model--direction--control comparisons, both when source OC and SJ labels
are shuffled and when fitted vectors are replaced with random directions of
comparable norm (Supplementary Fig.~S2). The main pattern remains robust across
confidence thresholds and a second Qwen2.5-7B strict-pair draw. A scoring-rule
audit likewise found no changes to SJ labels or the primary pair sample.

\section{Discussion}

\paragraph{Semantic validity of transferable readouts.}
A readout's transferability and semantic validity are distinct. When OC and SJ
conflict, conventional correctness-labelled contrasts often follow the model's
judgement rather than the ordering predicted by OC. After factorial
decomposition, the SJ-associated contrast predicts SJ above chance in all eight
held-out within-domain evaluations and preserves its expected polarity in all
eight model-by-direction cross-domain evaluations. The OC-associated
contrast does not preserve its polarity under cross-domain transfer and is
often reversed.
Transferability alone therefore does not establish objective-correctness
semantics.

\paragraph{Interpreting the transferable SJ signal.}
OC is externally defined by the relation between an answer and task-specific
ground truth. SJ is generated from cues available to the model, which may include
answer evidence, familiarity, fluency, or response policy. Math requires
multi-step numerical reasoning whereas Movies requires person-name recall, yet
the SJ-associated directions transfer between the domains despite their
different knowledge demands and error structures. At the final answer token,
the separation is weak in early
layers but sustained from middle to late layers, consistent with response-level
evaluation information becoming more separable at greater network depth. This
pattern may reflect endorsement or commitment as well as correctness
monitoring; the judgement measure does not distinguish among these accounts.

\paragraph{Implications for truth and metacognitive readouts.}
Aggregate correctness prediction does not identify what a probe reads out.
Semantic validation requires cases where external correctness and the model's
evaluation disagree, because aligned examples cannot separate them. Before the
separate judgement prompt, the final answer-token representation predicts the
model's later behavioural judgement without establishing privileged introspective
access
\cite{song2025failIntrospect,singh2026introspectionRealityCheck}.

\paragraph{Scope.}
The diagnostic sample comprises high-confidence judgements for questions
yielding usable correct and incorrect responses; it does not estimate
ordinary-output prevalence. Evidence spans four instruction-tuned models up to
14B and two free-response domains; OOD uses externally specified answer-letter
candidates. The factorial contrasts characterize associations between linear activation-mean directions
and subsequent behavioural judgements; they do not assess nonlinear or task-adapted correctness predictors.
Broader source-side elicitation, larger models, free-response OOD tasks, and causal interventions along fitted
directions would clarify generality and mechanism.

Across tested models and diagnostic subsets, the most transferable component is
more closely associated with subsequent self-judgement than externally scored
correctness.

\bibliography{references}

\clearpage
\onecolumn
\setcounter{section}{0}
\setcounter{subsection}{0}
\setcounter{figure}{0}
\setcounter{table}{0}
\setcounter{equation}{0}
\setcounter{secnumdepth}{2}
\renewcommand{\thesection}{S\arabic{section}}
\renewcommand{\thesubsection}{S\arabic{section}.\arabic{subsection}}
\renewcommand{\thefigure}{S\arabic{figure}}
\renewcommand{\thetable}{S\arabic{table}}
\renewcommand{\theequation}{S\arabic{equation}}
\captionsetup{font=small,labelfont=bf}

\begin{center}
{\LARGE\bfseries Supplementary Material}
\end{center}

\section{Supplementary Overview}

This appendix documents the data construction, representation contrasts, statistical procedures, and robustness analyses underlying the main paper. Table~\ref{tab:roadmap} links each experimental claim to its supporting analyses. Unless stated otherwise, bootstrap intervals are obtained by resampling target questions.

\begin{table}[h!]
\caption{Roadmap from the main experiments to supplementary evidence.}
\label{tab:roadmap}
\centering
\small
\begin{tabularx}{\textwidth}{lXX}
\toprule
Analysis & Main question & Supplementary material \\
\midrule
Exp1 & Does a correctness-labelled mixed direction follow OC or SJ on B/C conflicts? & Strict sample flow and the OC-only contrast (Secs.~S4 and S8). \\
Exp2A & Do the factorial contrasts preserve their intended rankings on held-out questions within domain? & Grouped cross-validation curves and fixed-window estimates (Sec.~S7). \\
Exp2B & Does the SJ component transfer more strongly than the OC component? & Formal transfer estimands, peaks, matching, question effects, null directions, answer likelihood, and construction robustness (Secs.~S5--S8). \\
OOD & Does transfer persist without fitting a direction on the target task? & MMLU and binary TruthfulQA construction, support, heterogeneity, and forced-choice controls (Sec.~S10). \\
\bottomrule
\end{tabularx}
\end{table}

\section{Datasets and Stimulus Construction}

\subsection{Math reasoning mixture}

The Math domain contains 16,700 problems drawn from eight locally defined source tags.
The pool includes GSM8K \cite{cobbe2021training}, ASDiv \cite{miao2021diverse}, and
SVAMP \cite{patel2021are}, together with MATH
\cite{hendrycks2021math}, MultiArith \cite{roy2016solving}, a MATH-500 subset
associated with process-supervision evaluation \cite{lightman2023lets}, and
two AMC subsets. The source tags indicate local provenance only and are not used as experimental strata.

A Qwen2.5-1.5B-Instruct pilot generated eight solutions per problem with
temperature 0.6 and a maximum generation length of 4,096 tokens. Its prompt
ended with: ``Please reason step by step, and put your final answer within
\texttt{\textbackslash boxed\{\}}.'' We retained items satisfying
$n_{\mathrm{correct}}<8$ and mean pilot output length $<380$ tokens. This
removes questions solved on every pilot attempt and the longest approximately
10\% of pilot responses while retaining all-wrong pilot questions. Model-specific
strict-pair construction later requires both a correct and an incorrect usable
answer. The resulting manuscript-facing pool contains 5,549 questions
(Table~\ref{tab:math-composition}).

All 5,549 references are numeric. Retained responses have one \texttt{Answer:}
field and normalize to a nonnegative integer or decimal; expressions and multiple candidate answers were excluded from the usable response pool. Correctness labels were obtained using normalized and symbolic-equivalence checks.

\begin{table}[tbp]
\caption{Math source composition before and after pilot filtering. Counts reflect the data files used in the reported analyses.}
\label{tab:math-composition}
\centering
\begin{tabular}{lrr}
\toprule
Source tag & Raw & Cleaned \\
\midrule
gsm8k & 8,702 & 3,394 \\
ASDiv & 2,248 & 907 \\
SVAMP & 1,000 & 524 \\
math & 3,073 & 377 \\
amc & 814 & 180 \\
MultiArith & 580 & 126 \\
math500 & 224 & 37 \\
amc2023 & 59 & 4 \\
Total & 16,700 & 5,549 \\
\bottomrule
\end{tabular}

\end{table}

\subsection{Movies factual recall}

The Movies domain contains 17,856 person-name recall records from the Movies
QA data of \citet{orgad2024llmsKnowMore}: 10,000 in the distributed train file
and 7,856 in the test file. Questions follow the form ``Who acted as [character]
in the movie [movie]?'' The answer instruction requests only the actor name.
Responses are lowercased, Unicode-normalized, stripped of diacritics, and
whitespace-normalized; a response is marked correct when the normalized
record-level reference actor appears in the normalized response. Unlike Math,
this domain requires direct entity recall rather than multi-step numerical
reasoning.

The 17,856 source records contain 17,518 unique question strings. The remaining repetitions arise from duplicate source entries and from missing or non-unique character fields, including blank roles and labels such as ``Himself'' or ``Herself''. Among the repeated prompts, 159 are associated with more than one reference actor. Depending on the model, prompts from this multi-reference set account for 1.4\%--2.4\% of retained rows. Identical prompt strings are assigned to the same question cluster in grouped cross-validation and bootstrap inference. A complementary sensitivity analysis removes the complete multi-reference set before direction fitting or evaluation (Sec.~S8.7). The sample-flow table accordingly reports both the number of retained response pairs and the number of unique question clusters.

\subsection{MMLU and binary TruthfulQA}
The out-of-distribution evaluation uses 400 MMLU test questions
\cite{hendrycks2021measuring}, sampled with a fixed seed as 100 questions from
each of the humanities, other, social-sciences, and STEM categories. Exact
subjects and question identifiers are retained in the released data artifact.
Each question is expanded into four candidate-response records, one for each answer letter, yielding 1,600 forced-choice responses. Binary TruthfulQA \cite{lin2022truthfulqa,lin2025truthfulqaData} comprises 790 questions.
Each question is paired with its designated true and false candidate answers,
yielding 1,580 candidate-response records.

In out-of-distribution tasks, each candidate answer letter is inserted as the assistant response and followed by the same Yes/No self-judgement query used in the main domains.
This procedure measures the model's judgement of correct and incorrect candidates at an identical,
exactly controlled response site. Neither MMLU nor TruthfulQA contributes data to direction fitting.

\section{Complete Prompts and Activation Site}

\subsection{Free-response prompts}

The complete Math user template is:
\begin{quote}\footnotesize\ttfamily\raggedright
Question: \{question\}\par
Strictly output your response in the following exact format:\par
Reasoning: A concise reasoning focusing on the core steps needed to solve the problem.\par
Answer: <final numeric answer only>.
\end{quote}

The complete Movies user template is:
\begin{quote}\footnotesize\ttfamily\raggedright
\{question\}\par
Return only the name of the actor.
\end{quote}

After a free-response answer was sampled, the judgement conversation was formed by
retaining the original user message and sampled assistant answer and appending the following user turn:
\begin{quote}\footnotesize\ttfamily\raggedright
Do you believe the answer above is correct? Answer only with Yes or No.
\end{quote}
Next-token Yes/No scores define the stored judgement probability $p_{\mathrm{judge}}$.
This judgement turn is used only to define self-judgement (SJ) and is excluded from the sequence
used for activation extraction.

\subsection{Forced-choice prompts}

For MMLU and binary TruthfulQA, the user message contains the question and all
answer options, followed by:
\begin{quote}\footnotesize\ttfamily\raggedright
Answer only with a single letter.
\end{quote}
The assistant turn is set to one candidate answer letter.
The same judgement turn shown above is then appended to compute $p_{\mathrm{judge}}$.
Candidate expansion represents every target question with both objectively correct and
objectively incorrect responses, without target-task answer generation or target-task direction fitting.

\subsection{Answer-token extraction}
For each retained response, the model-specific chat template is used to render the
user prompt and assistant answer. Before tokenization, preprocessing removes any terminal chat marker (for example, \texttt{<|im\_end|>}, \texttt{<|eot\_id|>}, or \texttt{<|endoftext|>}), and removes a terminal period when present.
The resulting sequence is tokenized, and the residual-stream hidden state at the output of each complete transformer block is recorded at the final non-padding sequence position. This is the block output rather than an intermediate attention or MLP activation. Accordingly, $x_{i,\ell}$ is aligned with the final numeric-answer token in Math, the final actor-name token in Movies, and the forced answer-letter token in the OOD tasks.
Forward passes are performed in bfloat16, and the resulting activation arrays are stored in float16.
Layer indices refer to transformer blocks and range from zero to $L-1$.

\begin{table}[tbp]
\caption{Prompt roles and stored signals. OC denotes objective correctness, and SJ denotes the thresholded high-confidence self-judgement label.}
\label{tab:prompt-summary}
\centering
\small
\begin{tabularx}{\textwidth}{lXX}
\toprule
Stage & Message content & Recorded variable \\
\midrule
Math answer & Reasoning plus a final numeric answer & Final answer-token hidden state; parsed OC \\
Movies answer & Actor name only & Final name-token hidden state; normalized-match OC \\
Self-judgement & Yes/No judgement of the preceding answer & $p_{\mathrm{judge}}$ and filtered SJ \\
OOD answer & Forced option letter & Final letter-token hidden state; option-key OC \\
\bottomrule
\end{tabularx}
\end{table}

\section{Strict-Pair Construction and Sample Flow}

For each model and main domain, strict pairs are produced as follows:
\begin{enumerate}
\item Generate eight stochastic answers per question. Main-model sampling uses
temperature 1.0 and maximum generation lengths of 384 tokens for Math and 256
for Movies.
\item Apply the domain-specific correctness parser and discard unusable
responses.
\item Retain questions with at least one correct and one incorrect usable
response, then sample one response of each type. Self-judgement is elicited for
this sampled pair, after which the paired confidence criterion is applied.
\item Compute $p_{\mathrm{judge}}$ for both selected responses using the same
model. The one-token call uses temperature 1.0 and retains up to four
next-token log-probability candidates. If the literal \texttt{Yes} and
\texttt{No} candidates are both present, their probabilities are normalized as
\[
p_{\mathrm{judge}}=
\frac{\exp(\ell_{\mathrm{Yes}})}
{\exp(\ell_{\mathrm{Yes}})+\exp(\ell_{\mathrm{No}})}.
\]
If only \texttt{Yes} is present, its full-vocabulary probability is used; if
only \texttt{No} is present, its complement is used; and if neither is present,
$p_{\mathrm{judge}}$ is set to 0.5. Rows assigned 0.5 by this final fallback are
excluded by the primary symmetric threshold at $\tau=0.7$.
\item Retain the pair only if both responses satisfy
$p_{\mathrm{judge}}\leq 1-\tau$ or $p_{\mathrm{judge}}\geq\tau$, with
$\tau=0.7$ in the primary analysis.
\item Assign $\mathrm{SJ}=0$ on the lower interval and $\mathrm{SJ}=1$ on the
upper interval. With tuples ordered as (OC,SJ), the four cells are
\[
A=(1,1),\quad B=(1,0),\quad C=(0,1),\quad D=(0,0).
\]
\end{enumerate}

Section~\ref{sec:construction-robustness} evaluates two upstream construction
choices directly. Full Yes/No normalization changes neither SJ labels nor
paired retention in any of the eight model--domain cells, while a second
Qwen2.5-7B draw from the frozen pass@8 pools preserves the cross-domain layer
profiles and positive fixed-window effects.

Each retained source item contributes one objectively correct and one
objectively incorrect response. Because SJ is measured rather than experimentally assigned,
the B and C cell counts need not be equal. Table~\ref{tab:id-attrition} reports the model-specific sample flow.
``Eligible pairs'' contain one sampled correct response and one sampled incorrect response before judgement filtering.
``Strict pairs'' additionally require both responses to satisfy the paired $\tau=0.7$ criterion.
``Unique $q$'' denotes exact prompt-string clusters and can therefore be smaller than the number of pairs
when multiple source records contain the same prompt.

\begin{table}[tbp]
\caption{Main-domain sample flow and strict-sample cell counts. Generated
items are source records presented to each model. The number of retained rows is
twice the number of strict pairs. Unique $q$ denotes exact prompt-string clusters
used for grouped cross-validation and bootstrap inference.}
\label{tab:id-attrition}
\centering
\scriptsize
\resizebox{\textwidth}{!}{\begin{tabular}{llrrrrrrrrrr}
\toprule
Model & Domain & Generated items & Eligible pairs & Strict pairs & Unique $q$ & Rows & A & B & C & D & B/C \\
\midrule
Qwen2.5-7B & Math & 5,549 & 2,703 & 2,307 & 2,307 & 4,614 & 2,139 & 168 & 1,719 & 588 & 1,887 \\
Qwen2.5-7B & Movies & 17,856 & 1,101 & 912 & 909 & 1,824 & 769 & 143 & 370 & 542 & 513 \\
Llama-3.1-8B & Math & 5,549 & 3,394 & 1,539 & 1,539 & 3,078 & 624 & 915 & 358 & 1,181 & 1,273 \\
Llama-3.1-8B & Movies & 17,856 & 3,682 & 2,780 & 2,775 & 5,560 & 927 & 1,853 & 293 & 2,487 & 2,146 \\
Qwen2.5-14B & Math & 5,549 & 2,605 & 2,370 & 2,370 & 4,740 & 1,056 & 1,314 & 249 & 2,121 & 1,563 \\
Qwen2.5-14B & Movies & 17,856 & 1,766 & 1,660 & 1,659 & 3,320 & 470 & 1,190 & 150 & 1,510 & 1,340 \\
OLMo-3-7B & Math & 5,549 & 1,584 & 362 & 362 & 724 & 258 & 104 & 221 & 141 & 325 \\
OLMo-3-7B & Movies & 17,856 & 1,093 & 508 & 508 & 1,016 & 375 & 133 & 116 & 392 & 249 \\
\bottomrule
\end{tabular}
}
\end{table}

The resulting estimand is conditional on questions for which the model produces both a correct response
and an incorrect response and assigns high-confidence judgements to the sampled pair.
The analyses therefore characterize the relation between OC and SJ within this diagnostic subset
rather than the prevalence of OC--SJ conflict cases in unconstrained model outputs.

\section{Factorial Contrasts and Formal Derivations}

\subsection{Cell means and cancellation of the interaction}

Let $o=2\mathrm{OC}-1$ and $s=2\mathrm{SJ}-1$. For domain $t$, consider the
associative representation model
\begin{equation}
x=\alpha_t o v^{\mathrm{OC}}_t+\beta_t s v^{\mathrm{SJ}}_t
+\eta_t os v^{\mathrm{INT}}_t+\epsilon,
\label{eq:assoc}
\end{equation}
where $v^{\mathrm{OC}}_t$, $v^{\mathrm{SJ}}_t$, and
$v^{\mathrm{INT}}_t$ are domain-specific directions and
$\alpha_t,\beta_t,\eta_t$ are their response-level effect magnitudes.
$\epsilon$ is a zero-mean residual term. Suppressing the domain subscript for readability,
the expected cell representations are
\begin{align}
\mu_A &= \alpha v^{\mathrm{OC}}+\beta v^{\mathrm{SJ}}+\eta v^{\mathrm{INT}},\\
\mu_B &= \alpha v^{\mathrm{OC}}-\beta v^{\mathrm{SJ}}-\eta v^{\mathrm{INT}},\\
\mu_C &=-\alpha v^{\mathrm{OC}}+\beta v^{\mathrm{SJ}}-\eta v^{\mathrm{INT}},\\
\mu_D &=-\alpha v^{\mathrm{OC}}-\beta v^{\mathrm{SJ}}+\eta v^{\mathrm{INT}}.
\end{align}
The two factorial main effects are
\begin{align}
W_{\mathrm{meta}}&=\frac{(\mu_A-\mu_B)+(\mu_C-\mu_D)}{2}
=2\beta v^{\mathrm{SJ}},\label{eq:wmeta}\\
W_{\mathrm{truth}}&=\frac{(\mu_A-\mu_C)+(\mu_B-\mu_D)}{2}
=2\alpha v^{\mathrm{OC}}.\label{eq:wtruth}
\end{align}
The interaction cancels in both contrasts. By comparison, the conventional
aligned contrast is
\begin{equation}
W_{\mathrm{mix}}=\mu_A-\mu_D
=2\alpha v^{\mathrm{OC}}+2\beta v^{\mathrm{SJ}}
=W_{\mathrm{truth}}+W_{\mathrm{meta}}.
\end{equation}
Thus, whenever both main-effect components are present, $W_{\mathrm{mix}}$ combines OC- and SJ-associated variation
rather than separating OC-associated variation from SJ-associated variation.

\subsection{Conflict-set ordering}

For a score $r_W(x)=\langle x,W\rangle$, define
\begin{equation}
\mathrm{AUC}_{C:B}(W)=\Pr\{r_W(X_C)>r_W(X_B)\},
\end{equation}
with ties assigned half weight. Cell B contains objectively correct but self-rejected responses,
whereas cell C contains objectively incorrect but self-endorsed responses.
An OC-associated score should rank B above C, yielding $\mathrm{AUC}_{C:B}<0.5$;
an SJ-associated score should rank C above B, yielding $\mathrm{AUC}_{C:B}>0.5$.
Within one domain, Eq.~\ref{eq:assoc}
gives
\begin{equation}
\mathbb{E}[r_{W_{\mathrm{mix}}}(X_C)-r_{W_{\mathrm{mix}}}(X_B)]
=4\beta^2\|v^{\mathrm{SJ}}\|^2-4\alpha^2\|v^{\mathrm{OC}}\|^2.
\end{equation}
The ordering of B and C under $W_{\mathrm{mix}}$ therefore indicates whether
the SJ- or OC-associated component contributes more strongly to the mixed contrast in Exp1.

\subsection{Cross-domain transfer}

Let $W^a_{\mathrm{meta}}$ and $W^a_{\mathrm{truth}}$ denote directions estimated in source domain $a$,
and let $x^b$ denote an activation from target domain $b$. The corresponding transfer scores are
\begin{equation}
r^{a\rightarrow b}_{\mathrm{meta}}(x^b)=
\langle x^b,W^a_{\mathrm{meta}}\rangle,
\qquad
r^{a\rightarrow b}_{\mathrm{truth}}(x^b)=
\langle x^b,W^a_{\mathrm{truth}}\rangle,
\end{equation}
where both directions are estimated from the source-domain cell means. Their scale does not affect the AUC rankings. Transfer performance is summarized by
$\mathrm{AUC}(r_{\mathrm{meta}}\rightarrow\mathrm{SJ})$ and
$\mathrm{AUC}(r_{\mathrm{truth}}\rightarrow\mathrm{OC})$.

Within the B/C conflict set, OC and SJ induce complementary class assignments.
Their relative transfer asymmetry is
\begin{equation}
\Delta_{\mathrm{CB}}=
\mathrm{AUC}(W_{\mathrm{meta}}\rightarrow\mathrm{SJ})-
\mathrm{AUC}(W_{\mathrm{truth}}\rightarrow\mathrm{OC}).
\label{eq:deltacb}
\end{equation}
The two component AUCs indicate whether each source direction transfers above chance,
whereas $\Delta_{\mathrm{CB}}$ compares their transfer performance on the same target responses.

The target-domain mean difference is
$\mu_C^b-\mu_B^b=-2\alpha_bv_b^{\mathrm{OC}}+2\beta_bv_b^{\mathrm{SJ}}$.
Its projection onto a source SJ direction increases with cross-domain alignment between
$v_a^{\mathrm{SJ}}$ and $v_b^{\mathrm{SJ}}$ and decreases with alignment between
the source SJ direction and the target OC component.
Exp2B estimates these relations without imposing orthogonality between the latent components.

\subsection{Source-question-adjusted direction sensitivity}

The pooled factorial directions weight the four source cells equally. A
complementary estimator removes additive source-question variation before
estimating the component directions. Let $x_q^+$ and $x_q^-$ denote the
activations of the correct and incorrect responses retained for source question
$q$, and let $s_q^+,s_q^-\in\{-1,+1\}$ denote their signed SJ labels. Under the
response-level model
\begin{equation}
x_q^{\pm}=h_q+b_{\mathrm{OC}}o_q^{\pm}
+b_{\mathrm{SJ}}s_q^{\pm}
+b_{\mathrm{INT}}o_q^{\pm}s_q^{\pm}+\epsilon_q^{\pm},
\end{equation}
where $o_q^+=+1$ and $o_q^-=-1$, the within-question difference is
\begin{equation}
d_q=x_q^+-x_q^-=2b_{\mathrm{OC}}
+(s_q^+-s_q^-)b_{\mathrm{SJ}}
+(s_q^++s_q^-)b_{\mathrm{INT}}+\widetilde\epsilon_q.
\label{eq:source-pair-fe}
\end{equation}
The additive question intercept $h_q$ cancels exactly. At each layer, multivariate
least squares across source pairs estimates the three coefficient vectors;
$\widehat b_{\mathrm{SJ}}$ and $\widehat b_{\mathrm{OC}}$ replace
$W_{\mathrm{meta}}$ and $W_{\mathrm{truth}}$, respectively, in the unchanged
cross-domain B/C evaluation. The four possible source-pair patterns are A/C,
A/D, B/C, and B/D, with design rows
$[2,s_q^+-s_q^-,s_q^++s_q^-]$. Identifiability is checked from the rank of
this three-column design for every source sample and bootstrap replicate.

\subsection{Target-question-controlled all-cell model}

The B/C analysis focuses on responses for which OC and SJ conflict.
A complementary analysis uses responses from all four cells.
For each component $k\in\{\mathrm{meta},\mathrm{truth}\}$,
projection scores are standardized separately within each target-domain and layer combination
and then averaged over the fixed layer window. The resulting scores are entered into
\begin{equation}
z^{(k)}_{iq}=\gamma_q+\beta^{(k)}_{\mathrm{OC}}o_{iq}
+\beta^{(k)}_{\mathrm{SJ}}s_{iq}
+\beta^{(k)}_{\mathrm{INT}}o_{iq}s_{iq}+\epsilon_{iq}.
\label{eq:fe}
\end{equation}
The coefficients are therefore identified from variation among responses associated with the same question.
The cross-component contrast is
\begin{equation}
\Delta_{\mathrm{FE}}=
\beta^{(\mathrm{meta})}_{\mathrm{SJ}}-
\beta^{(\mathrm{truth})}_{\mathrm{OC}}.
\end{equation}
As a within-component semantic diagnostic, we also compute
\begin{align}
\Delta^{(\mathrm{meta})}_{\mathrm{spec}}
&=\beta^{(\mathrm{meta})}_{\mathrm{SJ}}-
\beta^{(\mathrm{meta})}_{\mathrm{OC}},\\
\Delta^{(\mathrm{truth})}_{\mathrm{spec}}
&=\beta^{(\mathrm{truth})}_{\mathrm{OC}}-
\beta^{(\mathrm{truth})}_{\mathrm{SJ}}.
\end{align}
A positive value denotes the specificity implied by the source construction:
SJ for $W_{\mathrm{meta}}$ and OC for $W_{\mathrm{truth}}$.
Each question-cluster bootstrap replicate repeats both score standardization and model estimation.

\subsection{Response-likelihood residualization}

For free responses, response log-probability is the mean teacher-forced
log-probability over all tokens in the generated assistant response. Within each target dataset and
layer, each projection score is residualized against mean response log-probability
and response-token count. OOD residual models additionally include answer-letter
indicators. Component AUCs are computed from the layer-specific residual scores
and then averaged over the fixed window, matching the primary estimand.
Layer-wise nuisance regressions are refitted in every bootstrap replicate.

\section{Evaluation and Statistical Inference}

\paragraph{Direction fitting and centering.}
At each layer, cell means and the source centering vector are estimated exclusively from
the source-domain rows or the training partition. AUCs are computed from the resulting projection scores;
their rankings are invariant to positive rescaling of the fitted directions.
For the all-cell fixed-effect analysis, the two projected components are standardized separately before their coefficients are compared.

Exp2A uses five-fold \texttt{GroupKFold}, with the question-cluster identifier as the grouping variable,
so that the correct and incorrect responses associated with the same question remain in the same fold.
Out-of-fold scores are concatenated before evaluation. Exp2B estimates each direction once from
all strict source-domain rows and uses no target-domain labels during direction estimation.

\paragraph{Layer summaries.}
Layer-wise curves describe how each effect varies across network depth.
Primary fixed-window summaries average layers satisfying
$\ell/(L-1)\in[0.40,0.80]$: layers 11--21 for the 28-layer Qwen2.5-7B,
13--24 for the 32-layer models, and 19--37 for Qwen2.5-14B. Within a bootstrap
replicate, AUCs are computed by layer and then averaged over this fixed window.

The window covers a broad middle-to-late portion of each architecture, from 40\% to 80\% of normalized depth,
rather than selecting a single layer from the observed results.
Complete layer-wise curves are reported alongside the fixed-window summaries.
For descriptive peak analyses, the model-level peak is the layer that
maximizes the mean $\Delta_{\mathrm{CB}}$ across the two Math--Movies transfer directions.
Peak intervals are descriptive and are not interpreted as selection-adjusted hypothesis tests.

\paragraph{Cluster bootstrap.}
Unless noted otherwise, confidence intervals are based on 1,000 question-cluster bootstrap resamples.
Sampling a question retains all associated response rows, thereby preserving
the B/C pairing and the repeated candidate responses in the OOD tasks.
For the primary Exp2B fixed-window analysis, source and target question clusters are resampled independently,
and both factorial directions are re-estimated in every replicate.
For the all-layer figures and secondary analyses, the fitted source direction is held fixed,
so the resulting intervals quantify target-population uncertainty conditional on that direction.
Percentile 95\% intervals are computed from valid replicates.
All-layer counts summarize rows of the layer-by-direction result grid
and are descriptive because adjacent layers are not independent observations.

\begin{table}[tbp]
\caption{Descriptive all-layer summaries. ``CI above'' counts layer-direction
rows whose target-question 95\% interval is above the corresponding reference
(0.5 for Exp1, zero for the remaining analyses).}
\label{tab:all-layer-counts}
\centering
\small
\begin{tabular}{lcc}
\toprule
Analysis & Point estimate above reference & CI above reference \\
\midrule
Exp1 cross-domain $W_{\mathrm{mix}}$ & 256/280 (91.4\%) & 192/280 (68.6\%) \\
Exp2A in-domain $\Delta_{\mathrm{CB}}$ & 222/280 (79.3\%) & 176/280 (62.9\%) \\
Exp2B cross-domain $\Delta_{\mathrm{CB}}$ & 255/280 (91.1\%) & 192/280 (68.6\%) \\
OOD $\Delta_{\mathrm{CB}}$ & 458/560 (81.8\%) & 318/560 (56.8\%) \\
\bottomrule
\end{tabular}
\end{table}

\section{In-Domain Validation and Layer-Wise Results}

\begin{figure}[tbp]
\centering
\includegraphics[width=\textwidth]{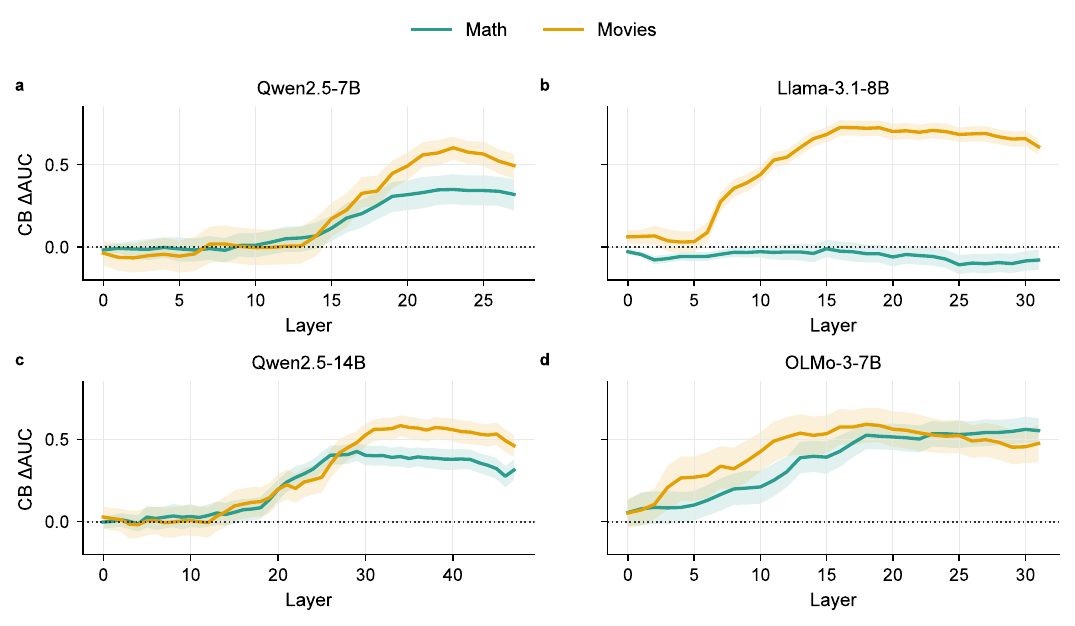}
\caption{Question-grouped in-domain validation (Exp2A). Each direction is fitted
within the training questions of a five-fold split and evaluated on held-out
questions. Curves show $\Delta_{\mathrm{CB}}$ by layer; shaded bands are 95\%
target-question cluster-bootstrap intervals. The pale vertical band marks the
fixed normalized layer window $[0.40,0.80]$.}
\label{fig:exp2a-supp}
\end{figure}

Figure~\ref{fig:exp2a-supp} reports Exp2A with question-grouped folds. Seven of the eight fixed-window estimates are positive, with 95\% confidence intervals excluding zero (Table~\ref{tab:exp2a-window}).
The only estimate whose interval includes zero is Llama-3.1-8B on Math,
$\Delta_{\mathrm{CB}}=-0.042$ [$-0.097$, $0.012$], while the same model on
Movies shows the largest in-domain contrast. Across models and domains,
positive contrasts generally emerge after the earliest layers and
remain present across much of the fixed middle-to-late layer window.

\begin{table}[tbp]
\caption{Exp2A fixed-window component AUCs and relative transfer asymmetry. Intervals use
1,000 question-cluster bootstrap resamples of the out-of-fold predictions.}
\label{tab:exp2a-window}
\centering
\scriptsize
\resizebox{\textwidth}{!}{\begin{tabular}{llcccc}
\toprule
Model & Domain & $W_{\mathrm{meta}}\!\rightarrow$SJ & $W_{\mathrm{truth}}\!\rightarrow$OC & $\Delta_{\mathrm{CB}}$ & $n_{B/C}$ \\
\midrule
Qwen2.5-7B & Math & 0.649 [0.607, 0.692] & 0.476 [0.436, 0.519] & 0.173 [0.096, 0.251] & 1,887 \\
Qwen2.5-7B & Movies & 0.723 [0.687, 0.763] & 0.483 [0.429, 0.543] & 0.240 [0.150, 0.329] & 513 \\
Llama-3.1-8B & Math & 0.652 [0.622, 0.680] & 0.693 [0.662, 0.723] & -0.042 [-0.097, 0.012] & 1,273 \\
Llama-3.1-8B & Movies & 0.915 [0.897, 0.931] & 0.219 [0.188, 0.250] & 0.696 [0.650, 0.742] & 2,146 \\
Qwen2.5-14B & Math & 0.743 [0.715, 0.766] & 0.395 [0.365, 0.430] & 0.348 [0.288, 0.398] & 1,563 \\
Qwen2.5-14B & Movies & 0.758 [0.725, 0.792] & 0.352 [0.311, 0.393] & 0.406 [0.336, 0.474] & 1,340 \\
OLMo-3-7B & Math & 0.899 [0.865, 0.928] & 0.422 [0.360, 0.484] & 0.476 [0.395, 0.556] & 325 \\
OLMo-3-7B & Movies & 0.867 [0.817, 0.916] & 0.315 [0.249, 0.381] & 0.552 [0.442, 0.660] & 249 \\
\bottomrule
\end{tabular}
}
\end{table}

Table~\ref{tab:exp2b-peaks} summarizes the descriptive peak layers for Exp2B. Within each model,
a single layer is selected by maximizing the mean point estimate of $\Delta_{\mathrm{CB}}$
across the two Math--Movies transfer directions. The resulting peak summaries describe the location of the
strongest observed cross-domain effect but are not used for the primary fixed-window inference.

\begin{table}[tbp]
\caption{Descriptive cross-domain peak layers. Confidence intervals condition
on the selected source direction and are not adjusted for peak selection.}
\label{tab:exp2b-peaks}
\centering
\small
\begin{tabular}{llccc}
\toprule
Model & Direction & Layer & Normalized depth & $\Delta_{CB}$ [95\% CI] \\
\midrule
Qwen2.5-7B & Math$\rightarrow$Movies & 19 & 0.704 & 0.672 [0.600, 0.739] \\
Qwen2.5-7B & Movies$\rightarrow$Math & 19 & 0.704 & 0.476 [0.383, 0.563] \\
Llama-3.1-8B & Math$\rightarrow$Movies & 19 & 0.613 & 0.617 [0.563, 0.677] \\
Llama-3.1-8B & Movies$\rightarrow$Math & 19 & 0.613 & 0.270 [0.202, 0.336] \\
Qwen2.5-14B & Math$\rightarrow$Movies & 28 & 0.596 & 0.523 [0.457, 0.593] \\
Qwen2.5-14B & Movies$\rightarrow$Math & 28 & 0.596 & 0.704 [0.660, 0.743] \\
OLMo-3-7B & Math$\rightarrow$Movies & 27 & 0.871 & 0.863 [0.809, 0.917] \\
OLMo-3-7B & Movies$\rightarrow$Math & 27 & 0.871 & 0.429 [0.315, 0.536] \\
\bottomrule
\end{tabular}

\end{table}

\subsection{Joint source--target bootstrap}

Table~\ref{tab:joint-bootstrap} reports the fixed-window inference used in the
main Exp2B table. Each replicate independently resamples source and target
question clusters, reconstructs $W_{\mathrm{meta}}$ and
$W_{\mathrm{truth}}$ at every selected layer, and evaluates the refitted
directions on the resampled target-domain B/C conflicts.
All 1,000 replicates were valid in each of the eight transfer conditions,
and every resampled source dataset retained observations from all four factorial cells.

\begin{table}[tbp]
\caption{Exp2B joint source--target question bootstrap. Source $q$ and target
$q$ are the numbers of question clusters entering direction fitting and B/C
evaluation, respectively. Values are fixed-window AUCs or differences with
95\% percentile intervals.}
\label{tab:joint-bootstrap}
\centering
\scriptsize
\resizebox{\textwidth}{!}{\begin{tabular}{llccccrr}
\toprule
Model & Direction & $W_{\mathrm{meta}}\!\rightarrow$SJ & $W_{\mathrm{truth}}\!\rightarrow$OC & $\Delta_{\mathrm{CB}}$ & Source $q$ & Target $q$ & Valid \\
\midrule
Qwen2.5-7B & Math$\rightarrow$Movies & 0.750 [0.715, 0.782] & 0.300 [0.263, 0.339] & 0.450 [0.386, 0.513] & 2,307 & 468 & 1,000 \\
Qwen2.5-7B & Movies$\rightarrow$Math & 0.670 [0.629, 0.708] & 0.368 [0.332, 0.414] & 0.302 [0.219, 0.371] & 909 & 1,824 & 1,000 \\
Llama-3.1-8B & Math$\rightarrow$Movies & 0.823 [0.796, 0.848] & 0.222 [0.191, 0.252] & 0.601 [0.545, 0.655] & 1,539 & 2,035 & 1,000 \\
Llama-3.1-8B & Movies$\rightarrow$Math & 0.572 [0.540, 0.603] & 0.434 [0.400, 0.471] & 0.138 [0.073, 0.198] & 2,775 & 1,240 & 1,000 \\
Qwen2.5-14B & Math$\rightarrow$Movies & 0.708 [0.679, 0.736] & 0.286 [0.259, 0.321] & 0.422 [0.365, 0.470] & 2,370 & 1,274 & 1,000 \\
Qwen2.5-14B & Movies$\rightarrow$Math & 0.781 [0.753, 0.802] & 0.215 [0.196, 0.246] & 0.566 [0.513, 0.602] & 1,659 & 1,493 & 1,000 \\
OLMo-3-7B & Math$\rightarrow$Movies & 0.928 [0.899, 0.950] & 0.167 [0.119, 0.243] & 0.761 [0.667, 0.818] & 362 & 249 & 1,000 \\
OLMo-3-7B & Movies$\rightarrow$Math & 0.776 [0.726, 0.821] & 0.441 [0.379, 0.511] & 0.334 [0.229, 0.430] & 508 & 318 & 1,000 \\
\bottomrule
\end{tabular}
}
\end{table}

\section{Main-Domain Controls}

\subsection{Source-question-adjusted direction sensitivity}

Table~\ref{tab:source-question-fe} reports the cross-domain evaluation after
estimating both source directions from Eq.~\ref{eq:source-pair-fe}. All eight
$\Delta_{\mathrm{CB}}$ estimates remain positive, ranging from 0.130 to 0.748,
and every joint source--target bootstrap 95\% interval excludes zero. The point
estimates differ from the corresponding pooled factorial estimates by at most
0.026. All eight source designs have rank three; all 1,000 bootstrap replicates
remain identifiable, including OLMo-3-7B Movies, where the A/C, A/D, and B/D
patterns identify the coefficients without observed B/C source pairs.

\begin{center}
\captionof{table}{Source-question-adjusted Exp2B sensitivity. Directions are estimated
from within-question correct-minus-incorrect activation differences. AC/AD/BC/BD
gives the source-pair support for the four SJ combinations; Rank is the
three-column design rank, and Valid is the number of identifiable joint
source--target bootstrap replicates. Entries are fixed-window AUCs or
differences with 95\% percentile intervals. The source-adjusted directions
$\widehat b_{\mathrm{SJ}}$ and $\widehat b_{\mathrm{OC}}$ occupy the evaluation
roles of the pooled $W_{\mathrm{meta}}$ and $W_{\mathrm{truth}}$, respectively.}
\label{tab:source-question-fe}
\scriptsize
\resizebox{\textwidth}{!}{\begin{tabular}{llccccrr}
\toprule
Model & Direction & $\widehat b_{\mathrm{SJ}}\!\rightarrow$SJ & $\widehat b_{\mathrm{OC}}\!\rightarrow$OC & $\Delta_{\mathrm{CB}}$ & AC/AD/BC/BD & Rank & Valid \\
\midrule
Qwen2.5-7B & Math$\rightarrow$Movies & 0.744 [0.709, 0.778] & 0.288 [0.253, 0.330] & 0.456 [0.389, 0.517] & 1,656/483/63/105 & 3 & 1,000 \\
Qwen2.5-7B & Movies$\rightarrow$Math & 0.676 [0.634, 0.714] & 0.379 [0.340, 0.423] & 0.297 [0.215, 0.366] & 325/444/45/98 & 3 & 1,000 \\
Llama-3.1-8B & Math$\rightarrow$Movies & 0.787 [0.754, 0.818] & 0.212 [0.183, 0.243] & 0.575 [0.514, 0.634] & 325/299/33/882 & 3 & 1,000 \\
Llama-3.1-8B & Movies$\rightarrow$Math & 0.573 [0.540, 0.604] & 0.443 [0.409, 0.477] & 0.130 [0.066, 0.191] & 185/742/108/1,745 & 3 & 1,000 \\
Qwen2.5-14B & Math$\rightarrow$Movies & 0.722 [0.690, 0.750] & 0.283 [0.256, 0.318] & 0.439 [0.375, 0.489] & 179/877/70/1,244 & 3 & 1,000 \\
Qwen2.5-14B & Movies$\rightarrow$Math & 0.785 [0.756, 0.806] & 0.224 [0.204, 0.257] & 0.562 [0.503, 0.596] & 85/385/65/1,125 & 3 & 1,000 \\
OLMo-3-7B & Math$\rightarrow$Movies & 0.896 [0.853, 0.928] & 0.148 [0.110, 0.213] & 0.748 [0.652, 0.810] & 214/44/7/97 & 3 & 1,000 \\
OLMo-3-7B & Movies$\rightarrow$Math & 0.746 [0.692, 0.794] & 0.425 [0.356, 0.504] & 0.321 [0.206, 0.420] & 116/259/0/133 & 3 & 1,000 \\
\bottomrule
\end{tabular}
}
\end{center}

\subsection{OC-only mass-mean control}

The OC-only direction is
$W_{\mathrm{OC-only}}=E[x\mid\mathrm{OC}=1]-E[x\mid\mathrm{OC}=0]$,
estimated from all source rows without SJ labels. This is a canonical
unwhitened mass-mean direction; because each retained source item contributes one
correct and one incorrect response, it is also equal to the average
within-question correct-minus-incorrect activation difference.

The fitted direction is evaluated on B/C conflicts in the target domain.
In all eight transfer conditions, it assigns higher scores to C responses,
which are incorrect but self-endorsed, than to B responses, which are correct but self-rejected (Table~\ref{tab:oc-only}).
Thus, the conflict-set reversal observed for the mixed contrast is also present
when the source direction is estimated solely from objective-correctness labels.

\begin{center}
\captionof{table}{Fixed-window OC-only mass-mean control. The direction is the
unwhitened source class-mean contrast
$E[x\mid\mathrm{OC}=1]-E[x\mid\mathrm{OC}=0]$, fitted without SJ labels.
AUC(C$>$B) above 0.5 means that this correctness-labelled source direction
ranks incorrect/self-endorsed target responses above correct/self-rejected
responses. Intervals condition on the fitted source direction.}
\label{tab:oc-only}
\small
\begin{tabular}{llccc}
\toprule
Model & Direction & AUC(C$>$B) [95\% CI] & Rows & Questions \\
\midrule
Qwen2.5-7B & Math$\rightarrow$Movies & 0.722 [0.688, 0.754] & 513 & 468 \\
Qwen2.5-7B & Movies$\rightarrow$Math & 0.653 [0.615, 0.691] & 1,887 & 1,824 \\
Llama-3.1-8B & Math$\rightarrow$Movies & 0.794 [0.763, 0.822] & 2,146 & 2,035 \\
Llama-3.1-8B & Movies$\rightarrow$Math & 0.567 [0.536, 0.599] & 1,273 & 1,240 \\
Qwen2.5-14B & Math$\rightarrow$Movies & 0.721 [0.691, 0.751] & 1,340 & 1,274 \\
Qwen2.5-14B & Movies$\rightarrow$Math & 0.788 [0.763, 0.811] & 1,563 & 1,493 \\
OLMo-3-7B & Math$\rightarrow$Movies & 0.874 [0.832, 0.911] & 249 & 249 \\
OLMo-3-7B & Movies$\rightarrow$Math & 0.686 [0.635, 0.733] & 325 & 318 \\
\bottomrule
\end{tabular}

\end{center}

\subsection{Window-level null directions}

\begin{figure}[tbp]
\centering
\includegraphics[width=\textwidth]{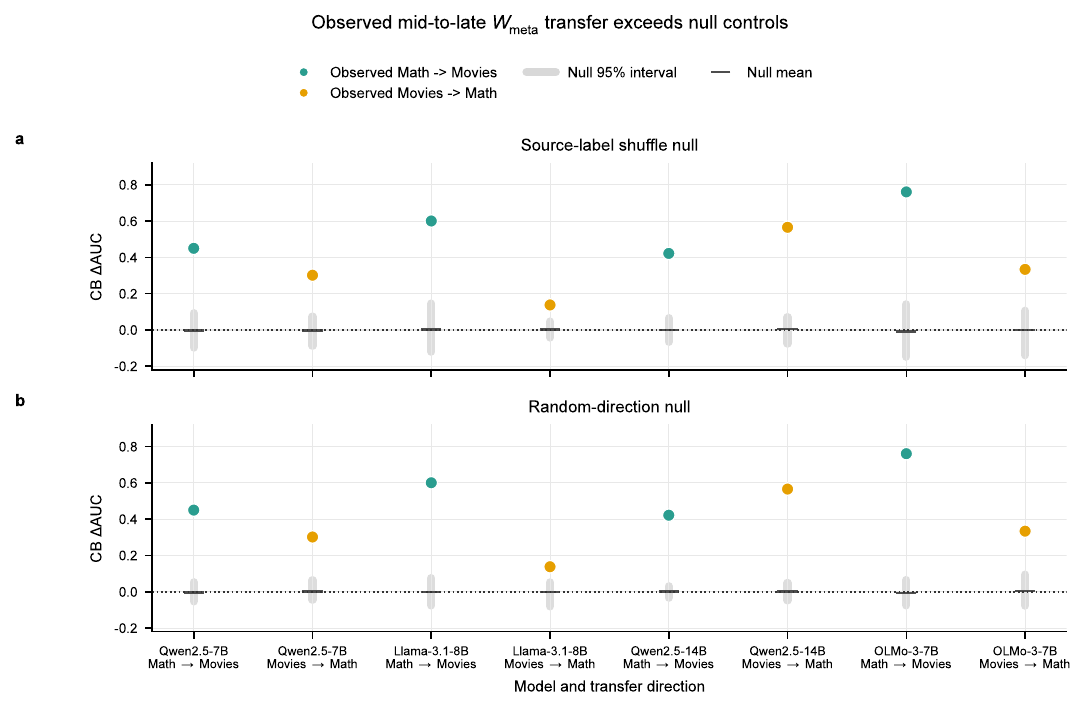}
\caption{Window-level null controls for Exp2B. Points show the observed
$\Delta_{\mathrm{CB}}$; null distributions are obtained from 100 source-label
shuffles or random direction pairs. The same target B/C rows and fixed
normalized layer window are used for observed and null estimates.}
\label{fig:nulls}
\end{figure}

Two nulls test whether large fixed-window contrasts arise from the target
quadrant structure alone. The source-label-shuffle null independently permutes
source OC/SJ labels before direction construction. The random-direction null
replaces each fitted direction with a normalized random vector of the same
dimensionality. Each null uses 100 repetitions and is evaluated on the unchanged
target conflict set. Figure~\ref{fig:nulls} and Table~\ref{tab:nulls} compare
their distributions with the observed fixed-window effects.

\begin{table}[tbp]
\caption{Numerical window-level null summaries. The empirical probability is
the plus-one Monte Carlo estimate $(b+1)/(R+1)$, where $b$ is the number of
valid null repetitions at least as large as the observed effect.}
\label{tab:nulls}
\centering
\scriptsize
\resizebox{\textwidth}{!}{\begin{tabular}{lllccc}
\toprule
Model & Direction & Null & Observed & Null mean [95\% range] & $p_{\mathrm{emp}}$ \\
\midrule
Llama-3.1-8B & Math$\rightarrow$Movies & Random direction & 0.601 & -0.000 [-0.075, 0.075] & 0.010 \\
Llama-3.1-8B & Math$\rightarrow$Movies & Label shuffle & 0.601 & 0.002 [-0.119, 0.145] & 0.010 \\
Llama-3.1-8B & Movies$\rightarrow$Math & Random direction & 0.138 & -0.001 [-0.080, 0.052] & 0.010 \\
Llama-3.1-8B & Movies$\rightarrow$Math & Label shuffle & 0.138 & 0.003 [-0.041, 0.046] & 0.010 \\
OLMo-3-7B & Math$\rightarrow$Movies & Random direction & 0.761 & -0.006 [-0.075, 0.065] & 0.010 \\
OLMo-3-7B & Math$\rightarrow$Movies & Label shuffle & 0.761 & -0.009 [-0.147, 0.141] & 0.010 \\
OLMo-3-7B & Movies$\rightarrow$Math & Random direction & 0.334 & 0.004 [-0.076, 0.094] & 0.010 \\
OLMo-3-7B & Movies$\rightarrow$Math & Label shuffle & 0.334 & -0.000 [-0.139, 0.106] & 0.010 \\
Qwen2.5-14B & Math$\rightarrow$Movies & Random direction & 0.422 & 0.000 [-0.032, 0.030] & 0.010 \\
Qwen2.5-14B & Math$\rightarrow$Movies & Label shuffle & 0.422 & 0.001 [-0.065, 0.065] & 0.010 \\
Qwen2.5-14B & Movies$\rightarrow$Math & Random direction & 0.566 & 0.000 [-0.047, 0.049] & 0.010 \\
Qwen2.5-14B & Movies$\rightarrow$Math & Label shuffle & 0.566 & 0.004 [-0.075, 0.071] & 0.010 \\
Qwen2.5-7B & Math$\rightarrow$Movies & Random direction & 0.450 & -0.003 [-0.053, 0.052] & 0.010 \\
Qwen2.5-7B & Math$\rightarrow$Movies & Label shuffle & 0.450 & -0.003 [-0.097, 0.092] & 0.010 \\
Qwen2.5-7B & Movies$\rightarrow$Math & Random direction & 0.302 & 0.003 [-0.044, 0.064] & 0.010 \\
Qwen2.5-7B & Movies$\rightarrow$Math & Label shuffle & 0.302 & -0.004 [-0.087, 0.074] & 0.010 \\
\bottomrule
\end{tabular}
}
\end{table}

\subsection{B/C token-count matching}

Token-count matching evaluates whether differences in complete rendered
question-plus-response sequence length account for the separation between B and C
responses. Matching is performed within the target domain before AUC computation
and does not alter source-direction estimation. The balance-optimized procedure evaluates alternative token-count coarsenings and selects the specification whose token-count-only SJ AUC is closest to 0.5, breaking ties in favor of retaining more responses. Standardized mean difference (SMD) is reported as an additional balance diagnostic. A second specification uses fixed pooled token-count deciles.

Table~\ref{tab:matching} reports the number of retained responses, post-matching balance,
and the resulting fixed-window contrasts. The estimated contrasts remain positive under both matching procedures,
including conditions in which achieving token-count balance requires substantial trimming.

\begin{table}[tbp]
\caption{Token-count matching diagnostics. SMD is the standardized B/C mean
difference in target rendered-sequence token count. Matching is fixed before inference;
intervals resample target questions within the matched set.}
\label{tab:matching}
\centering
\scriptsize
\resizebox{\textwidth}{!}{\begin{tabular}{lllrrrrc}
\toprule
Model & Direction & Rule & Rows & Keep & SMD pre & SMD post & $\Delta_{\mathrm{CB}}$ [95\% CI] \\
\midrule
Llama-3.1-8B & Math$\rightarrow$Movies & Optimized & 586 & 27.3\% & 0.104 & 0.012 & 0.625 [0.552, 0.684] \\
Llama-3.1-8B & Math$\rightarrow$Movies & Fixed deciles & 586 & 27.3\% & 0.104 & 0.027 & 0.622 [0.563, 0.684] \\
Llama-3.1-8B & Movies$\rightarrow$Math & Optimized & 468 & 36.8\% & -1.222 & 0.000 & 0.415 [0.333, 0.499] \\
Llama-3.1-8B & Movies$\rightarrow$Math & Fixed deciles & 480 & 37.7\% & -1.222 & -0.015 & 0.394 [0.318, 0.483] \\
OLMo-3-7B & Math$\rightarrow$Movies & Optimized & 202 & 81.1\% & 0.182 & 0.018 & 0.772 [0.697, 0.838] \\
OLMo-3-7B & Math$\rightarrow$Movies & Fixed deciles & 202 & 81.1\% & 0.182 & 0.018 & 0.772 [0.700, 0.837] \\
OLMo-3-7B & Movies$\rightarrow$Math & Optimized & 154 & 47.4\% & -1.079 & -0.075 & 0.198 [0.065, 0.329] \\
OLMo-3-7B & Movies$\rightarrow$Math & Fixed deciles & 154 & 47.4\% & -1.079 & -0.075 & 0.198 [0.066, 0.330] \\
Qwen2.5-14B & Math$\rightarrow$Movies & Optimized & 300 & 22.4\% & 0.139 & 0.027 & 0.418 [0.342, 0.486] \\
Qwen2.5-14B & Math$\rightarrow$Movies & Fixed deciles & 300 & 22.4\% & 0.139 & 0.055 & 0.488 [0.422, 0.559] \\
Qwen2.5-14B & Movies$\rightarrow$Math & Optimized & 498 & 31.9\% & -0.403 & 0.005 & 0.628 [0.569, 0.683] \\
Qwen2.5-14B & Movies$\rightarrow$Math & Fixed deciles & 498 & 31.9\% & -0.403 & 0.005 & 0.608 [0.542, 0.673] \\
Qwen2.5-7B & Math$\rightarrow$Movies & Optimized & 286 & 55.8\% & 0.129 & -0.016 & 0.429 [0.346, 0.509] \\
Qwen2.5-7B & Math$\rightarrow$Movies & Fixed deciles & 286 & 55.8\% & 0.129 & -0.016 & 0.429 [0.346, 0.505] \\
Qwen2.5-7B & Movies$\rightarrow$Math & Optimized & 336 & 17.8\% & -0.009 & -0.037 & 0.293 [0.198, 0.392] \\
Qwen2.5-7B & Movies$\rightarrow$Math & Fixed deciles & 336 & 17.8\% & -0.009 & -0.007 & 0.277 [0.174, 0.374] \\
\bottomrule
\end{tabular}
}
\end{table}

\subsection{Response likelihood}

Mean response log-probability itself often discriminates OC and SJ on the conflict
set, but it does not explain the projection asymmetry. Table~\ref{tab:main-logp}
reports raw and residualized effects after controlling each projection score
for mean response log-probability and response-token count at each layer. All eight residual
$\Delta_{\mathrm{CB}}$ estimates remain positive with 95\% intervals above
zero.

\begin{table}[tbp]
\caption{Free-response likelihood control. ``logp$\rightarrow$OC/SJ''
is the AUC of mean assistant-response token log-probability alone on the same B/C rows.
Residual effects control projection scores for mean response log-probability and
response-token count separately at each layer, followed by the same layer-AUC averaging as the
primary analysis. Nuisance models are refitted in every bootstrap replicate.}
\label{tab:main-logp}
\centering
\scriptsize
\resizebox{\textwidth}{!}{\begin{tabular}{llcccc}
\toprule
Model & Direction & logp$\rightarrow$OC & logp$\rightarrow$SJ & Raw $\Delta_{CB}$ & Residual $\Delta_{CB}$ \\
\midrule
Qwen2.5-7B & Math$\rightarrow$Movies & 0.692 & 0.308 & 0.450 [0.389, 0.513] & 0.470 [0.409, 0.529] \\
Qwen2.5-7B & Movies$\rightarrow$Math & 0.417 & 0.583 & 0.302 [0.227, 0.381] & 0.339 [0.275, 0.407] \\
Llama-3.1-8B & Math$\rightarrow$Movies & 0.645 & 0.355 & 0.601 [0.543, 0.651] & 0.624 [0.564, 0.673] \\
Llama-3.1-8B & Movies$\rightarrow$Math & 0.487 & 0.513 & 0.138 [0.076, 0.201] & 0.280 [0.219, 0.345] \\
Qwen2.5-14B & Math$\rightarrow$Movies & 0.562 & 0.438 & 0.422 [0.368, 0.477] & 0.431 [0.377, 0.485] \\
Qwen2.5-14B & Movies$\rightarrow$Math & 0.420 & 0.580 & 0.566 [0.520, 0.610] & 0.619 [0.575, 0.660] \\
OLMo-3-7B & Math$\rightarrow$Movies & 0.546 & 0.454 & 0.761 [0.694, 0.821] & 0.751 [0.678, 0.814] \\
OLMo-3-7B & Movies$\rightarrow$Math & 0.232 & 0.768 & 0.334 [0.236, 0.419] & 0.180 [0.083, 0.267] \\
\bottomrule
\end{tabular}
}
\end{table}

\subsection{Question fixed effects}

The all-cell model in Eq.~\ref{eq:fe} compares responses within question and
does not require restricting the fit to B/C rows. Across all eight transfers,
$\beta^{(\mathrm{meta})}_{\mathrm{SJ}}$ exceeds
$\beta^{(\mathrm{truth})}_{\mathrm{OC}}$ and the cluster interval for
$\Delta_{\mathrm{FE}}$ remains above zero (Table~\ref{tab:question-fe}). This
result links the conflict-set AUC pattern to the full $2\times2$ design.

The complete coefficient matrix in Table~\ref{tab:question-fe-full} shows that
both transferred components have positive OC and SJ cross-loadings. In every
direction, the SJ coefficient exceeds the OC coefficient for both
$W_{\mathrm{meta}}$ and $W_{\mathrm{truth}}$. The latter pattern is consistent
with the conflict-set reversal: a source correctness-associated contrast need
not retain OC specificity after transfer. The factorial names identify how the
source directions are constructed, not pure target-domain latent axes.
The within-component contrasts in Table~\ref{tab:question-fe-specificity}
quantify this asymmetry. All eight
$\Delta^{(\mathrm{meta})}_{\mathrm{spec}}$ estimates are positive with 95\%
intervals above zero (0.192--0.619), whereas all eight
$\Delta^{(\mathrm{truth})}_{\mathrm{spec}}$ estimates are negative with
intervals below zero ($-0.490$--$-0.214$). Thus the transferred
$W_{\mathrm{meta}}$ direction retains SJ specificity, while the transferred
$W_{\mathrm{truth}}$ direction loads more strongly on SJ than OC.

\begin{table}[!tbp]
\caption{Question-fixed-effect cross-domain projection analysis. Scores are
standardized by component and target layer, then averaged over the fixed
window.}
\label{tab:question-fe}
\centering
\small
\resizebox{0.8\textwidth}{!}{\begin{tabular}{llcccc}
\toprule
Model & Direction & $\beta^{meta}_{SJ}$ & $\beta^{truth}_{OC}$ & $\Delta_{FE}$ [95\% CI] & Questions \\
\midrule
Qwen2.5-7B & Math$\rightarrow$Movies & 0.515 & 0.173 & 0.342 [0.266, 0.419] & 909 \\
Qwen2.5-7B & Movies$\rightarrow$Math & 0.496 & 0.165 & 0.331 [0.270, 0.388] & 2,307 \\
Llama-3.1-8B & Math$\rightarrow$Movies & 0.615 & 0.211 & 0.403 [0.353, 0.455] & 2,775 \\
Llama-3.1-8B & Movies$\rightarrow$Math & 0.456 & 0.226 & 0.230 [0.155, 0.295] & 1,539 \\
Qwen2.5-14B & Math$\rightarrow$Movies & 0.414 & 0.113 & 0.301 [0.241, 0.358] & 1,659 \\
Qwen2.5-14B & Movies$\rightarrow$Math & 0.539 & 0.120 & 0.419 [0.374, 0.462] & 2,370 \\
OLMo-3-7B & Math$\rightarrow$Movies & 0.724 & 0.194 & 0.530 [0.432, 0.621] & 508 \\
OLMo-3-7B & Movies$\rightarrow$Math & 0.612 & 0.098 & 0.514 [0.314, 0.737] & 362 \\
\bottomrule
\end{tabular}
}
\end{table}

\begin{table}[!thbp]
\caption{Complete question-fixed-effect cross-loading matrix. Panel A projects
the source $W_{\mathrm{meta}}$ direction; Panel B projects the source
$W_{\mathrm{truth}}$ direction. Entries are coefficients with target-question
cluster-bootstrap 95\% confidence intervals.}
\label{tab:question-fe-full}
\centering
\scriptsize
\textit{Panel A: $W_{\mathrm{meta}}$ projection}\par\smallskip
\resizebox{\textwidth}{!}{\begin{tabular}{llccc}
\toprule
Model & Direction & $\beta_{OC}$ & $\beta_{SJ}$ & $\beta_{INT}$ \\
\midrule
Qwen2.5-7B & Math$\rightarrow$Movies & 0.172 [0.131, 0.211] & 0.515 [0.468, 0.560] & -0.124 [-0.174, -0.073] \\
Qwen2.5-7B & Movies$\rightarrow$Math & 0.200 [0.171, 0.232] & 0.496 [0.456, 0.535] & 0.135 [0.102, 0.166] \\
Llama-3.1-8B & Math$\rightarrow$Movies & 0.231 [0.211, 0.251] & 0.615 [0.580, 0.651] & -0.103 [-0.127, -0.081] \\
Llama-3.1-8B & Movies$\rightarrow$Math & 0.265 [0.241, 0.288] & 0.456 [0.399, 0.511] & -0.091 [-0.113, -0.068] \\
Qwen2.5-14B & Math$\rightarrow$Movies & 0.184 [0.159, 0.210] & 0.414 [0.369, 0.456] & -0.127 [-0.155, -0.096] \\
Qwen2.5-14B & Movies$\rightarrow$Math & 0.162 [0.140, 0.185] & 0.539 [0.509, 0.567] & 0.029 [0.003, 0.056] \\
OLMo-3-7B & Math$\rightarrow$Movies & 0.105 [0.061, 0.151] & 0.724 [0.659, 0.781] & -0.055 [-0.102, -0.011] \\
OLMo-3-7B & Movies$\rightarrow$Math & 0.142 [0.086, 0.194] & 0.612 [0.436, 0.807] & -0.106 [-0.165, -0.044] \\
\bottomrule
\end{tabular}
}
\par\medskip
\textit{Panel B: $W_{\mathrm{truth}}$ projection}\par\smallskip
\resizebox{\textwidth}{!}{\begin{tabular}{llccc}
\toprule
Model & Direction & $\beta_{OC}$ & $\beta_{SJ}$ & $\beta_{INT}$ \\
\midrule
Qwen2.5-7B & Math$\rightarrow$Movies & 0.173 [0.134, 0.207] & 0.423 [0.377, 0.472] & -0.096 [-0.142, -0.048] \\
Qwen2.5-7B & Movies$\rightarrow$Math & 0.165 [0.136, 0.196] & 0.378 [0.338, 0.419] & 0.098 [0.066, 0.130] \\
Llama-3.1-8B & Math$\rightarrow$Movies & 0.211 [0.188, 0.236] & 0.554 [0.517, 0.593] & -0.085 [-0.111, -0.060] \\
Llama-3.1-8B & Movies$\rightarrow$Math & 0.226 [0.203, 0.251] & 0.462 [0.403, 0.522] & -0.089 [-0.112, -0.065] \\
Qwen2.5-14B & Math$\rightarrow$Movies & 0.113 [0.082, 0.145] & 0.347 [0.302, 0.390] & -0.063 [-0.098, -0.027] \\
Qwen2.5-14B & Movies$\rightarrow$Math & 0.120 [0.100, 0.140] & 0.460 [0.433, 0.488] & 0.001 [-0.023, 0.026] \\
OLMo-3-7B & Math$\rightarrow$Movies & 0.194 [0.151, 0.242] & 0.685 [0.618, 0.745] & -0.149 [-0.192, -0.100] \\
OLMo-3-7B & Movies$\rightarrow$Math & 0.098 [0.044, 0.153] & 0.397 [0.224, 0.587] & -0.083 [-0.147, -0.020] \\
\bottomrule
\end{tabular}
}
\end{table}

\begin{table}[!thbp]
\caption{Within-component specificity diagnostics for the all-cell model.
Positive values indicate the source-implied target specificity: SJ for
$W_{\mathrm{meta}}$ and OC for $W_{\mathrm{truth}}$. Entries are coefficient
differences with target-question cluster-bootstrap 95\% confidence intervals.}
\label{tab:question-fe-specificity}
\centering
\small
\resizebox{0.8\textwidth}{!}{\begin{tabular}{llcc}
\toprule
Model & Direction & $\beta^{meta}_{SJ}-\beta^{meta}_{OC}$ [95\% CI] & $\beta^{truth}_{OC}-\beta^{truth}_{SJ}$ [95\% CI] \\
\midrule
Qwen2.5-7B & Math$\rightarrow$Movies & 0.343 [0.265, 0.423] & -0.250 [-0.332, -0.177] \\
Qwen2.5-7B & Movies$\rightarrow$Math & 0.296 [0.232, 0.355] & -0.214 [-0.275, -0.150] \\
Llama-3.1-8B & Math$\rightarrow$Movies & 0.384 [0.338, 0.433] & -0.343 [-0.395, -0.292] \\
Llama-3.1-8B & Movies$\rightarrow$Math & 0.192 [0.119, 0.260] & -0.235 [-0.303, -0.161] \\
Qwen2.5-14B & Math$\rightarrow$Movies & 0.230 [0.172, 0.282] & -0.234 [-0.294, -0.169] \\
Qwen2.5-14B & Movies$\rightarrow$Math & 0.378 [0.328, 0.423] & -0.340 [-0.384, -0.296] \\
OLMo-3-7B & Math$\rightarrow$Movies & 0.619 [0.515, 0.713] & -0.490 [-0.586, -0.384] \\
OLMo-3-7B & Movies$\rightarrow$Math & 0.469 [0.269, 0.689] & -0.298 [-0.514, -0.105] \\
\bottomrule
\end{tabular}
}
\end{table}

\subsection{Judgement scoring and data-construction robustness}
\label{sec:construction-robustness}

The stored judgement score was originally computed from the returned
next-token candidates as described in Sec.~S4. We recomputed the exact literal
\texttt{Yes} and \texttt{No} logits at the first assistant-generation position
for every pre-filter pair row. Because inference backends need not reproduce
token logits bit-for-bit, the scoring-rule audit holds these recomputed logits
fixed and compares the historical top-four fallback with full binary
normalization,
\(
p_{\mathrm{judge}}=\operatorname{sigmoid}
(z_{\mathrm{Yes}}-z_{\mathrm{No}})
\).
Across all eight model--domain cells, the maximum row-wise difference was
$1.30\times10^{-6}$. The two scoring rules produced no SJ label changes at
0.5 and no complete-pair membership changes under the primary paired
$\tau=0.7$ criterion. The candidate-return truncation therefore did not
determine the analyzed strict samples.

\paragraph{Strict-pair resampling.}
The main analysis samples one correct and one incorrect response from each
question's usable pass@8 responses. To measure sensitivity to this draw, we
constructed a second Qwen2.5-7B pair sample from the frozen response pools
using seed 2027. Each correct and incorrect response was drawn uniformly from
its corresponding pool; the originally selected response remained eligible,
and repeated generation slots retained their empirical multiplicity. At least
one response changed in 2,441/2,703 Math pairs (90.3\%) and 645/1,101 Movies
pairs (58.6\%). Full Yes/No normalization followed by the paired $\tau=0.7$
criterion retained 2,289 Math pairs and 907 Movies pairs, with observations in
all four factorial cells.

All-layer $\Delta_{\mathrm{CB}}$ profiles from the resampled pairs correlated
with the original profiles at 0.996 for Math$\rightarrow$Movies and 0.991 for
Movies$\rightarrow$Math. In each direction, 23/28 layer-wise estimates were
positive. The fixed-window effects remained positive under 1,000 joint
source--target question bootstrap resamples with source-direction refitting
(Table~\ref{tab:strict-pair-r2}). Thus the cross-domain pattern is preserved
under an independently repeated draw from the finite pass@8 response pools.

\begin{center}
\captionof{table}{Qwen2.5-7B strict-pair resampling sensitivity. ``Original'' gives the
main-analysis window effect; R2 columns are computed from the second
pair draw. Entries are fixed-window AUCs or differences with 95\% percentile
intervals from 1,000 joint source--target question bootstrap resamples.}
\label{tab:strict-pair-r2}
\scriptsize
\resizebox{\textwidth}{!}{\begin{tabular}{lcccc}
\toprule
Direction & Original $\Delta_{\mathrm{CB}}$ & R2 $W_{\mathrm{meta}}\!\rightarrow$SJ & R2 $W_{\mathrm{truth}}\!\rightarrow$OC & R2 $\Delta_{\mathrm{CB}}$ \\
\midrule
Math$\rightarrow$Movies & 0.450 [0.386, 0.513] & 0.763 [0.728, 0.796] & 0.315 [0.278, 0.358] & 0.448 [0.374, 0.514] \\
Movies$\rightarrow$Math & 0.302 [0.219, 0.371] & 0.654 [0.614, 0.695] & 0.405 [0.362, 0.448] & 0.249 [0.168, 0.330] \\
\bottomrule
\end{tabular}
}
\end{center}

\paragraph{Multi-reference Movies prompts.}
The 159 Movies prompt strings linked to more than one reference actor define a
separate label-ambiguity sensitivity. We removed every retained response whose
exact prompt belonged to this set before recomputing Exp2B. For
Math$\rightarrow$Movies this changes the target conflict set; for
Movies$\rightarrow$Math it changes the source sample, and both factorial
directions are refitted. The filter removed 1.38\%--2.41\% of retained Movies
rows across models and 6--39 B/C responses.

All eight filtered fixed-window effects remained positive with 95\% intervals
above zero (Table~\ref{tab:movies-multireference}). Their all-layer profiles
correlated with the original profiles at $r\geq0.999$, and the largest absolute
change in a window effect was 0.030. The weakest original condition,
Llama-3.1-8B Movies$\rightarrow$Math, changed from 0.138 to 0.140
[0.075, 0.204]. The cross-domain result is therefore unchanged when
multi-reference Movies prompts are excluded rather than handled only through
question clustering.

\begin{center}
\captionof{table}{Movies multi-reference prompt sensitivity. All retained responses
from prompts associated with multiple reference actors are removed before
direction fitting or target evaluation. Filtered entries are fixed-window
$\Delta_{\mathrm{CB}}$ estimates with 95\% percentile intervals from 1,000
joint source--target question bootstrap resamples with direction refitting.}
\label{tab:movies-multireference}
\scriptsize
\resizebox{0.8\textwidth}{!}{\begin{tabular}{lrlcc}
\toprule
Model & Removed rows (\%) & Direction & Original $\Delta_{\mathrm{CB}}$ & Filtered $\Delta_{\mathrm{CB}}$ \\
\midrule
Qwen2.5-7B & 44 (2.41) & Math$\rightarrow$Movies & 0.450 & 0.444 [0.374, 0.508] \\
Qwen2.5-7B & 44 (2.41) & Movies$\rightarrow$Math & 0.302 & 0.303 [0.221, 0.375] \\
Llama-3.1-8B & 80 (1.44) & Math$\rightarrow$Movies & 0.601 & 0.631 [0.572, 0.685] \\
Llama-3.1-8B & 80 (1.44) & Movies$\rightarrow$Math & 0.138 & 0.140 [0.075, 0.204] \\
Qwen2.5-14B & 54 (1.63) & Math$\rightarrow$Movies & 0.422 & 0.419 [0.359, 0.469] \\
Qwen2.5-14B & 54 (1.63) & Movies$\rightarrow$Math & 0.566 & 0.566 [0.512, 0.603] \\
OLMo-3-7B & 14 (1.38) & Math$\rightarrow$Movies & 0.761 & 0.753 [0.661, 0.816] \\
OLMo-3-7B & 14 (1.38) & Movies$\rightarrow$Math & 0.334 & 0.335 [0.222, 0.435] \\
\bottomrule
\end{tabular}
}
\end{center}

\section{Self-Judgement Threshold Sensitivity}

The primary strict analysis uses $\tau=0.7$ at the pair level. A complementary
row-level sensitivity analysis instead reconstructs the pre-filter population by joining the
original activation files with a supplementary activation patch for rows that
had been removed at $\tau=0.7$. For each
$\tau\in\{0.5,0.6,0.7,0.8\}$, it removes the uncertain band
$1-\tau<p_{\mathrm{judge}}<\tau$ and assigns SJ from the retained side. This is
a row-filter sensitivity population; the primary sample instead requires both
members of a question pair to pass $\tau=0.7$.

\begin{figure}[tbp]
\centering
\includegraphics[width=\textwidth]{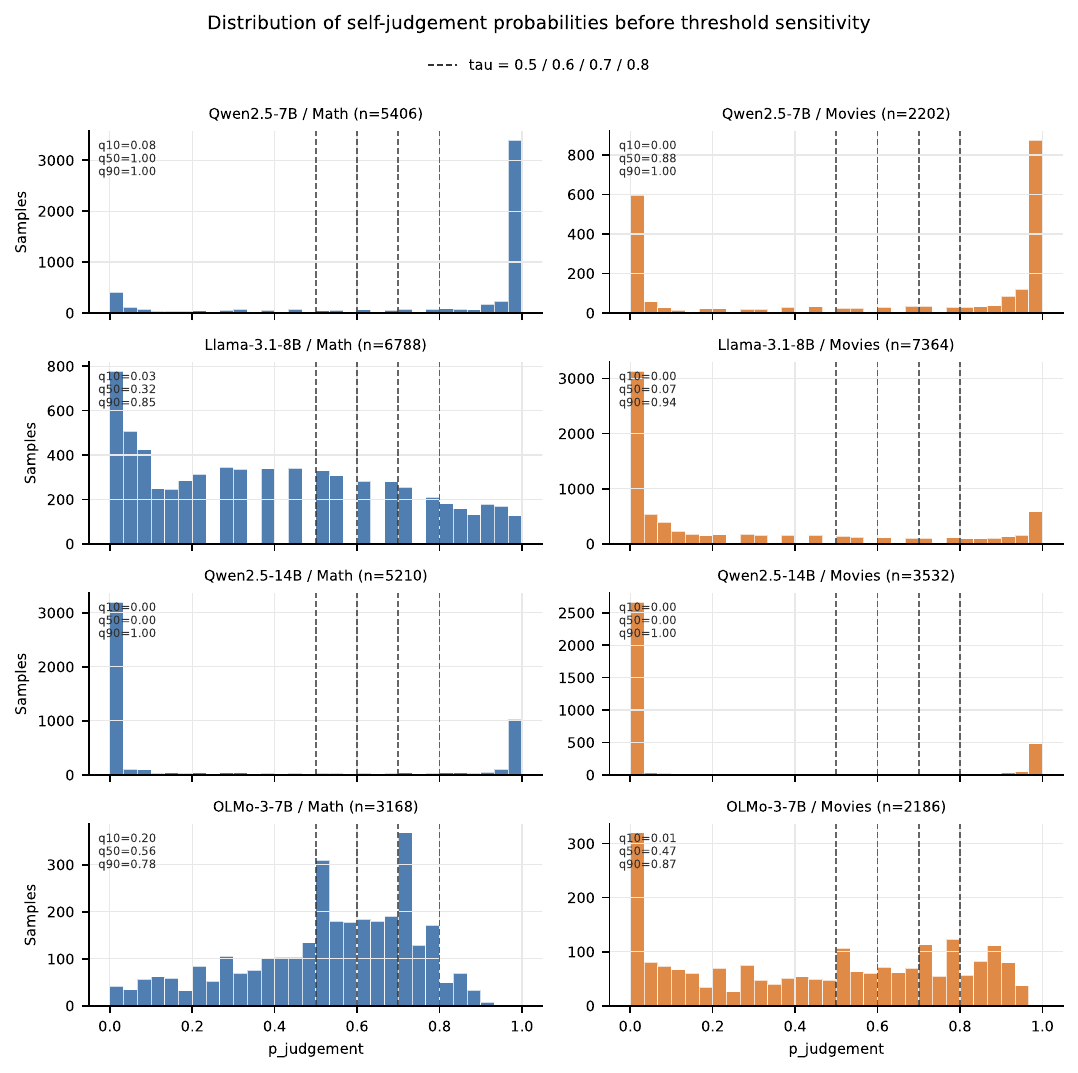}
\caption{Pre-filter distributions of $p_{\mathrm{judge}}$ for every model and
main domain. Vertical references mark the lower and upper boundaries induced by
the symmetric confidence thresholds. These probabilities precede the main
paired strict filter.}
\label{fig:threshold-dist}
\end{figure}

Figure~\ref{fig:threshold-dist} shows why threshold changes affect models
differently. Qwen distributions are highly polarized, whereas OLMo assigns much
more mass near 0.5. At $\tau=0.8$, only 14.0\% of OLMo Math rows remain, so
that setting is a support stress test rather than an equally powered analysis.
Table~\ref{tab:threshold-counts} gives the complete retained quadrant counts.

\begin{table}[!thbp]
\caption{Row-level support under symmetric confidence filtering. ``Kept'' is
the number of pre-filter judged rows outside the uncertain band.}
\label{tab:threshold-counts}
\centering
\scriptsize
\resizebox{\textwidth}{!}{\begin{tabular}{llcrrrrrrr}
\toprule
Model & Domain & $\tau$ & Kept & Keep & A & B & C & D & B/C \\
\midrule
Qwen2.5-7B & Math & 0.500 & 5,406 & 100.0\% & 2,459 & 244 & 1,942 & 761 & 2,186 \\
Qwen2.5-7B & Math & 0.600 & 5,230 & 96.7\% & 2,434 & 222 & 1,861 & 713 & 2,083 \\
Qwen2.5-7B & Math & 0.700 & 4,992 & 92.3\% & 2,408 & 194 & 1,769 & 621 & 1,963 \\
Qwen2.5-7B & Math & 0.800 & 4,751 & 87.9\% & 2,371 & 164 & 1,659 & 557 & 1,823 \\
Qwen2.5-7B & Movies & 0.500 & 2,202 & 100.0\% & 902 & 199 & 454 & 647 & 653 \\
Qwen2.5-7B & Movies & 0.600 & 2,122 & 96.4\% & 875 & 184 & 433 & 630 & 617 \\
Qwen2.5-7B & Movies & 0.700 & 2,007 & 91.1\% & 838 & 162 & 407 & 600 & 569 \\
Qwen2.5-7B & Movies & 0.800 & 1,903 & 86.4\% & 803 & 144 & 378 & 578 & 522 \\
Llama-3.1-8B & Math & 0.500 & 6,788 & 100.0\% & 1,707 & 1,687 & 909 & 2,485 & 2,596 \\
Llama-3.1-8B & Math & 0.600 & 5,809 & 85.6\% & 1,316 & 1,482 & 663 & 2,348 & 2,145 \\
Llama-3.1-8B & Math & 0.700 & 4,569 & 67.3\% & 964 & 1,096 & 451 & 2,058 & 1,547 \\
Llama-3.1-8B & Math & 0.800 & 3,440 & 50.7\% & 654 & 769 & 294 & 1,723 & 1,063 \\
Llama-3.1-8B & Movies & 0.500 & 7,364 & 100.0\% & 1,365 & 2,317 & 527 & 3,155 & 2,844 \\
Llama-3.1-8B & Movies & 0.600 & 6,923 & 94.0\% & 1,174 & 2,218 & 438 & 3,093 & 2,656 \\
Llama-3.1-8B & Movies & 0.700 & 6,371 & 86.5\% & 1,029 & 1,999 & 358 & 2,985 & 2,357 \\
Llama-3.1-8B & Movies & 0.800 & 5,789 & 78.6\% & 869 & 1,771 & 295 & 2,854 & 2,066 \\
Qwen2.5-14B & Math & 0.500 & 5,210 & 100.0\% & 1,193 & 1,412 & 316 & 2,289 & 1,728 \\
Qwen2.5-14B & Math & 0.600 & 5,109 & 98.1\% & 1,153 & 1,390 & 286 & 2,280 & 1,676 \\
Qwen2.5-14B & Math & 0.700 & 4,968 & 95.4\% & 1,110 & 1,349 & 261 & 2,248 & 1,610 \\
Qwen2.5-14B & Math & 0.800 & 4,816 & 92.4\% & 1,069 & 1,310 & 232 & 2,205 & 1,542 \\
Qwen2.5-14B & Movies & 0.500 & 3,532 & 100.0\% & 529 & 1,237 & 170 & 1,596 & 1,407 \\
Qwen2.5-14B & Movies & 0.600 & 3,486 & 98.7\% & 506 & 1,228 & 162 & 1,590 & 1,390 \\
Qwen2.5-14B & Movies & 0.700 & 3,424 & 96.9\% & 483 & 1,211 & 154 & 1,576 & 1,365 \\
Qwen2.5-14B & Movies & 0.800 & 3,376 & 95.6\% & 462 & 1,204 & 139 & 1,571 & 1,343 \\
OLMo-3-7B & Math & 0.500 & 3,168 & 100.0\% & 1,153 & 431 & 898 & 686 & 1,329 \\
OLMo-3-7B & Math & 0.600 & 2,159 & 68.2\% & 809 & 276 & 575 & 499 & 851 \\
OLMo-3-7B & Math & 0.700 & 1,357 & 42.8\% & 507 & 184 & 322 & 344 & 506 \\
OLMo-3-7B & Math & 0.800 & 445 & 14.0\% & 98 & 98 & 62 & 187 & 160 \\
OLMo-3-7B & Movies & 0.500 & 2,186 & 100.0\% & 783 & 310 & 308 & 785 & 618 \\
OLMo-3-7B & Movies & 0.600 & 1,806 & 82.6\% & 642 & 234 & 219 & 711 & 453 \\
OLMo-3-7B & Movies & 0.700 & 1,465 & 67.0\% & 511 & 163 & 148 & 643 & 311 \\
OLMo-3-7B & Movies & 0.800 & 1,004 & 45.9\% & 301 & 103 & 67 & 533 & 170 \\
\bottomrule
\end{tabular}
}
\end{table}

The sensitivity run uses 100 question-cluster bootstrap resamples. Across all
280 Exp2B layer-direction rows, positive $\Delta_{\mathrm{CB}}$ occurs in
228, 239, 252, and 251 rows at $\tau=0.5,0.6,0.7,$ and 0.8, respectively;
the corresponding intervals lie above zero in 178, 193, 196, and 205 rows.
The cross-domain pattern therefore persists across alternative confidence
filters. Increasing $\tau$ removes more rows near the decision boundary and
generally strengthens separation, while reducing support most sharply for
OLMo.

\section{Zero-Target-Fitting OOD Analyses}

Source directions are fitted only on Math or Movies strict samples and applied
unchanged to MMLU and binary TruthfulQA. OOD records retain only high-confidence
B/C candidates under the same symmetric $\tau=0.7$ rule. Because several
candidates can originate from one question, all intervals cluster by the
original target question.

Table~\ref{tab:ood-heterogeneity} reports support and the full layer-by-source
descriptive counts. Positive $\Delta_{\mathrm{CB}}$ appears in 458/560
(81.8\%) rows, with question-cluster intervals above zero in 318/560 (56.8\%).
The main OOD figure's component curves show that this difference is driven by
$W_{\mathrm{meta}}\rightarrow\mathrm{SJ}$ rising above chance while
$W_{\mathrm{truth}}\rightarrow\mathrm{OC}$ remains lower on the same target
conflicts. The result occurs for both a four-option knowledge benchmark and a
binary truthfulness benchmark, although effect magnitude varies by model,
source, and target.

\begin{table}[!thbp]
\caption{OOD support and all-layer heterogeneity. ``Rows'' in the final two
columns are source-by-layer evaluations: two source directions per model-target
pair times the number of model layers.}
\label{tab:ood-heterogeneity}
\centering
\small
\resizebox{0.8\textwidth}{!}{\begin{tabular}{llrrrrcc}
\toprule
Model & Target & B & C & Questions & Mean $\Delta_{CB}$ & $\Delta>0$ & CI$>0$ \\
\midrule
Llama-3.1-8B & MMLU & 13 & 403 & 252 & 0.206 & 47/64 & 31/64 \\
Llama-3.1-8B & TruthfulQA & 357 & 160 & 459 & 0.267 & 53/64 & 46/64 \\
OLMo-3-7B & MMLU & 70 & 128 & 160 & 0.223 & 57/64 & 44/64 \\
OLMo-3-7B & TruthfulQA & 448 & 84 & 476 & 0.228 & 54/64 & 42/64 \\
Qwen2.5-14B & MMLU & 169 & 75 & 205 & 0.207 & 69/96 & 48/96 \\
Qwen2.5-14B & TruthfulQA & 383 & 48 & 388 & 0.357 & 91/96 & 66/96 \\
Qwen2.5-7B & MMLU & 112 & 189 & 224 & 0.072 & 43/56 & 12/56 \\
Qwen2.5-7B & TruthfulQA & 401 & 147 & 483 & 0.154 & 44/56 & 29/56 \\
\bottomrule
\end{tabular}
}
\end{table}

\subsection{Forced-choice likelihood and answer-letter controls}
Forced-choice answer likelihood is measured as the teacher-forced probability of the inserted option letter.
Within the B/C conflict set, its OC and SJ AUCs are complementary by construction,
but their direction and magnitude vary across models and target datasets.
To control for candidate likelihood and option identity, each layer-specific projection score is
residualized with respect to option log-probability, token count, and answer-letter indicators.
The nuisance regression is re-estimated within each bootstrap replicate.
Table~\ref{tab:ood-logp} reports the resulting contrast.
Fifteen of sixteen letter-controlled estimates remain positive with intervals
above zero. The exception is OLMo Movies$\rightarrow$TruthfulQA,
$\Delta_{\mathrm{CB}}=-0.030$ [-0.138, 0.077], whose 95\% CI includes zero.

\begin{table}[!thbp]
\caption{OOD answer-likelihood and letter control. Raw
$\Delta_{\mathrm{CB}}$ is the component difference before residualization;
the final column controls answer-option log-probability, target token count, and
answer letter. Binary TruthfulQA has one non-reference letter indicator; MMLU
has three.}
\label{tab:ood-logp}
\centering
\scriptsize
\resizebox{\textwidth}{!}{\begin{tabular}{llcccc}
\toprule
Model & Direction & logp$\rightarrow$OC & logp$\rightarrow$SJ & Raw $\Delta_{CB}$ & Letter+logp residual $\Delta_{CB}$ \\
\midrule
Qwen2.5-7B & Math$\rightarrow$MMLU & 0.670 & 0.330 & 0.112 & 0.185 [0.080, 0.280] \\
Qwen2.5-7B & Math$\rightarrow$TruthfulQA & 0.497 & 0.503 & 0.141 & 0.142 [0.051, 0.226] \\
Qwen2.5-7B & Movies$\rightarrow$MMLU & 0.670 & 0.330 & 0.091 & 0.200 [0.112, 0.283] \\
Qwen2.5-7B & Movies$\rightarrow$TruthfulQA & 0.497 & 0.503 & 0.227 & 0.242 [0.177, 0.305] \\
Llama-3.1-8B & Math$\rightarrow$MMLU & 0.557 & 0.443 & 0.350 & 0.513 [0.292, 0.725] \\
Llama-3.1-8B & Math$\rightarrow$TruthfulQA & 0.349 & 0.651 & 0.396 & 0.402 [0.322, 0.484] \\
Llama-3.1-8B & Movies$\rightarrow$MMLU & 0.557 & 0.443 & 0.333 & 0.627 [0.415, 0.811] \\
Llama-3.1-8B & Movies$\rightarrow$TruthfulQA & 0.349 & 0.651 & 0.425 & 0.455 [0.369, 0.532] \\
Qwen2.5-14B & Math$\rightarrow$MMLU & 0.704 & 0.296 & 0.389 & 0.395 [0.303, 0.476] \\
Qwen2.5-14B & Math$\rightarrow$TruthfulQA & 0.389 & 0.611 & 0.579 & 0.620 [0.533, 0.698] \\
Qwen2.5-14B & Movies$\rightarrow$MMLU & 0.704 & 0.296 & 0.342 & 0.398 [0.302, 0.482] \\
Qwen2.5-14B & Movies$\rightarrow$TruthfulQA & 0.389 & 0.611 & 0.581 & 0.566 [0.481, 0.649] \\
OLMo-3-7B & Math$\rightarrow$MMLU & 0.511 & 0.489 & 0.458 & 0.452 [0.323, 0.558] \\
OLMo-3-7B & Math$\rightarrow$TruthfulQA & 0.246 & 0.754 & 0.518 & 0.406 [0.317, 0.498] \\
OLMo-3-7B & Movies$\rightarrow$MMLU & 0.511 & 0.489 & 0.233 & 0.252 [0.103, 0.386] \\
OLMo-3-7B & Movies$\rightarrow$TruthfulQA & 0.246 & 0.754 & 0.193 & -0.030 [-0.138, 0.077] \\
\bottomrule
\end{tabular}
}
\end{table}

\subsection{Counterbalanced target-side judgement labels}

The primary SJ label uses a Yes/No response. To test whether the OOD result is
tied to assent or a fixed response-token mapping, we re-elicited target-side
judgements with two prompts:
\begin{quote}
\small
Evaluate whether the answer above is correct. Respond with X if it is correct
and Y if it is incorrect.

Evaluate whether the answer above is correct. Respond with Y if it is correct
and X if it is incorrect.
\end{quote}
X/Y avoids reusing the A--D labels already assigned to forced-choice answers.
For each candidate, we compute correctness-oriented log odds
\[
z_{XY}=\log P(X)-\log P(Y),\qquad
z_{YX}=\log P(Y)-\log P(X),
\]
and define
\[
z_{\mathrm{bal}}=\tfrac12(z_{XY}+z_{YX}),\qquad
p_{\mathrm{bal}}=\operatorname{sigmoid}(z_{\mathrm{bal}}).
\]
The arithmetic mean of the two mapping-corrected probabilities is stored only
as a secondary diagnostic and is not used to define the reported labels or
transfer results. The primary analysis applies the same symmetric
$\tau=0.7$ rule to $p_{\mathrm{bal}}$. Because this changes which candidates
belong to B and C, answer-token activations were extracted for the complete new
conflict set at the same forced-answer final-token site. Math- and Movies-fitted
directions, layer windows, and target-question bootstrap clusters remain
unchanged.

Mapping-corrected XY and YX scores correlate strongly in every model--target
cell ($\rho=0.917$--$0.975$), and their jointly high-confidence binary labels
agree in 92.7\%--100\% of candidates (Table~\ref{tab:ood-cb-diagnostics}). The
balanced MMLU conflict sets are sparse on B for Llama-3.1-8B ($n_B=13$) and
OLMo-3-7B ($n_B=4$); those cells consequently provide weak raw inferential
support despite having many C rows.

\begin{table}[!thbp]
\caption{Counterbalanced target-side judgement diagnostics. $\rho_{XY,YX}$ is
the Spearman correlation between the two mapping-corrected log odds. HC
agreement is computed where both mappings yield a high-confidence binary
label. Retained counts all candidates outside the symmetric uncertain band;
B/C and conflict questions describe the balanced conflict set.}
\label{tab:ood-cb-diagnostics}
\centering
\scriptsize
\resizebox{0.8\textwidth}{!}{\begin{tabular}{llcrrrrr}
\toprule
Model & Target & $\rho_{XY,YX}$ & HC agreement & Retained & B & C & Conflict $q$ \\
\midrule
Llama-3.1-8B & MMLU & 0.920 & 95.7\% & 1,112 & 13 & 454 & 258 \\
Llama-3.1-8B & TruthfulQA & 0.937 & 98.9\% & 1,125 & 283 & 196 & 420 \\
OLMo-3-7B & MMLU & 0.917 & 99.6\% & 1,361 & 4 & 857 & 369 \\
OLMo-3-7B & TruthfulQA & 0.975 & 100.0\% & 1,259 & 102 & 318 & 372 \\
Qwen2.5-14B & MMLU & 0.941 & 98.1\% & 1,581 & 135 & 140 & 201 \\
Qwen2.5-14B & TruthfulQA & 0.963 & 96.5\% & 1,550 & 270 & 94 & 292 \\
Qwen2.5-7B & MMLU & 0.943 & 92.7\% & 1,573 & 260 & 32 & 280 \\
Qwen2.5-7B & TruthfulQA & 0.949 & 93.7\% & 1,529 & 506 & 82 & 535 \\
\bottomrule
\end{tabular}
}
\end{table}

Across the 16 model--source--target rows, fixed-window
$\Delta_{\mathrm{CB}}$ is positive in all cases (mean 0.383, range
0.176--0.682), with 12 raw question-cluster intervals above zero
(Table~\ref{tab:ood-cb-transfer}). The four intervals crossing zero are exactly
the two source directions evaluated in each of the low-B MMLU cells. After
layer-wise adjustment for answer-option log-probability and token count, all 16
point estimates and intervals are above zero. Across individual layers,
464/560 (82.9\%) estimates are positive and 282/560 (50.4\%) intervals are
above zero. This control supports target-side judgement-label robustness;
the source directions were still constructed from the original elicitation.

\begin{table}[!thbp]
\caption{Fixed-window OOD transfer under counterbalanced X/Y judgement labels.
Component AUCs and raw $\Delta_{\mathrm{CB}}$ use the complete balanced B/C
sets. Residual $\Delta_{\mathrm{CB}}$ controls answer-option log-probability and
token count separately at each layer. All intervals use 1,000
target-question-cluster bootstrap resamples.}
\label{tab:ood-cb-transfer}
\centering
\scriptsize
\resizebox{\textwidth}{!}{\begin{tabular}{llcccc}
\toprule
Model & Direction & $W_{\mathrm{meta}}\!\rightarrow$SJ & $W_{\mathrm{truth}}\!\rightarrow$OC & Raw $\Delta_{CB}$ & Residual $\Delta_{CB}$ \\
\midrule
Qwen2.5-7B & Math$\rightarrow$MMLU & 0.590 [0.514, 0.661] & 0.390 [0.322, 0.467] & 0.200 [0.054, 0.332] & 0.220 [0.065, 0.353] \\
Qwen2.5-7B & Math$\rightarrow$TruthfulQA & 0.575 [0.517, 0.631] & 0.399 [0.343, 0.461] & 0.176 [0.056, 0.284] & 0.172 [0.053, 0.281] \\
Qwen2.5-7B & Movies$\rightarrow$MMLU & 0.642 [0.571, 0.715] & 0.424 [0.339, 0.500] & 0.218 [0.078, 0.373] & 0.301 [0.152, 0.449] \\
Qwen2.5-7B & Movies$\rightarrow$TruthfulQA & 0.632 [0.586, 0.672] & 0.388 [0.347, 0.432] & 0.244 [0.157, 0.321] & 0.234 [0.139, 0.318] \\
Llama-3.1-8B & Math$\rightarrow$MMLU & 0.692 [0.562, 0.808] & 0.386 [0.220, 0.582] & 0.305 [-0.015, 0.577] & 0.378 [0.089, 0.628] \\
Llama-3.1-8B & Math$\rightarrow$TruthfulQA & 0.707 [0.664, 0.753] & 0.311 [0.264, 0.355] & 0.396 [0.311, 0.486] & 0.335 [0.251, 0.414] \\
Llama-3.1-8B & Movies$\rightarrow$MMLU & 0.678 [0.502, 0.826] & 0.356 [0.192, 0.541] & 0.322 [-0.036, 0.633] & 0.481 [0.169, 0.763] \\
Llama-3.1-8B & Movies$\rightarrow$TruthfulQA & 0.723 [0.676, 0.767] & 0.288 [0.244, 0.335] & 0.435 [0.342, 0.521] & 0.375 [0.294, 0.459] \\
Qwen2.5-14B & Math$\rightarrow$MMLU & 0.599 [0.546, 0.654] & 0.367 [0.316, 0.418] & 0.231 [0.135, 0.336] & 0.257 [0.167, 0.348] \\
Qwen2.5-14B & Math$\rightarrow$TruthfulQA & 0.768 [0.713, 0.819] & 0.266 [0.214, 0.323] & 0.502 [0.392, 0.599] & 0.486 [0.376, 0.590] \\
Qwen2.5-14B & Movies$\rightarrow$MMLU & 0.649 [0.607, 0.691] & 0.459 [0.409, 0.507] & 0.190 [0.105, 0.278] & 0.280 [0.193, 0.361] \\
Qwen2.5-14B & Movies$\rightarrow$TruthfulQA & 0.810 [0.765, 0.847] & 0.201 [0.166, 0.243] & 0.609 [0.524, 0.681] & 0.590 [0.506, 0.665] \\
OLMo-3-7B & Math$\rightarrow$MMLU & 0.841 [0.504, 0.984] & 0.239 [0.018, 0.601] & 0.602 [-0.100, 0.963] & 0.671 [0.249, 0.976] \\
OLMo-3-7B & Math$\rightarrow$TruthfulQA & 0.858 [0.820, 0.892] & 0.176 [0.138, 0.216] & 0.682 [0.606, 0.751] & 0.629 [0.542, 0.699] \\
OLMo-3-7B & Movies$\rightarrow$MMLU & 0.788 [0.415, 0.986] & 0.276 [0.092, 0.593] & 0.513 [-0.178, 0.895] & 0.584 [0.185, 0.894] \\
OLMo-3-7B & Movies$\rightarrow$TruthfulQA & 0.795 [0.741, 0.844] & 0.291 [0.237, 0.350] & 0.504 [0.395, 0.605] & 0.429 [0.324, 0.525] \\
\bottomrule
\end{tabular}
}
\end{table}

\begin{figure}[thbp]
\centering
\includegraphics[width=\textwidth]{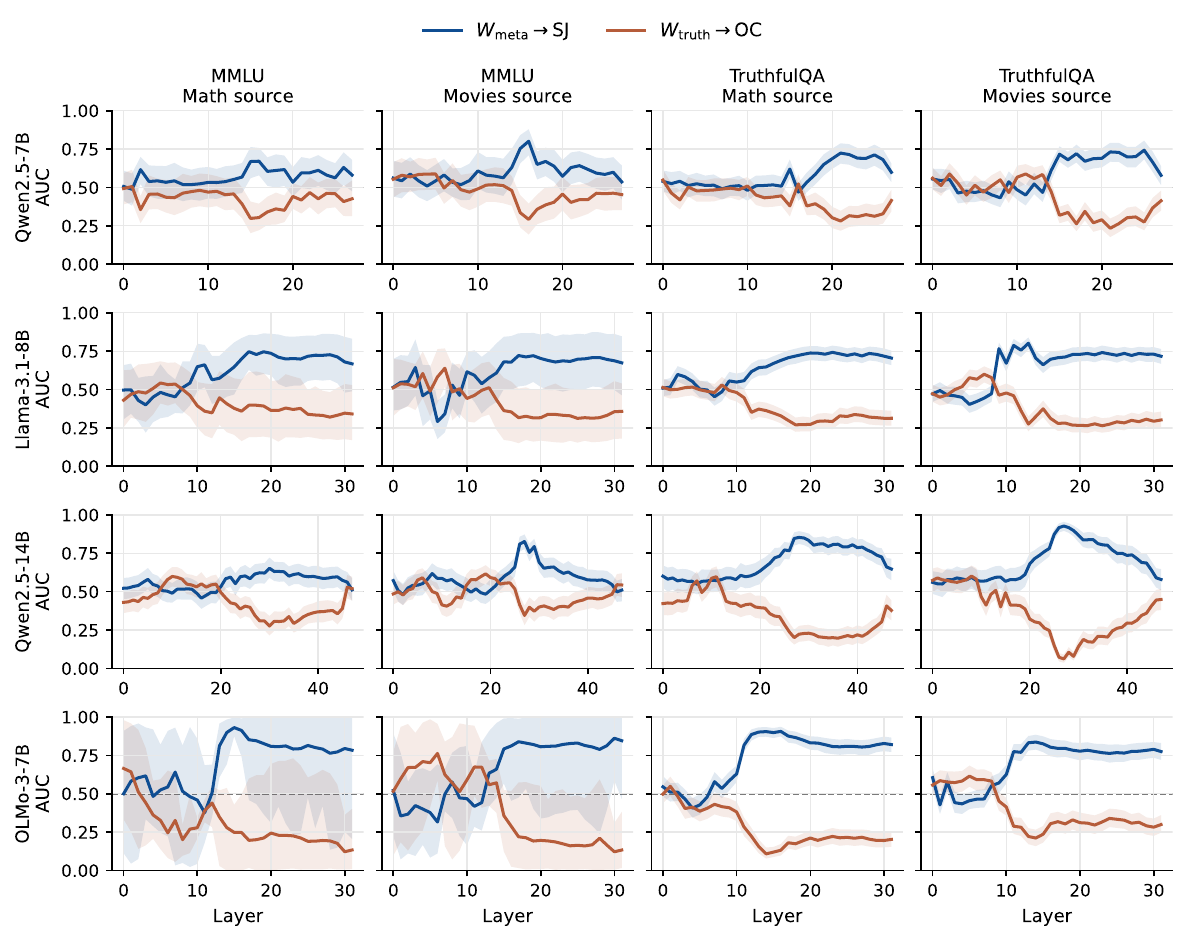}
\caption{All-layer component AUCs under counterbalanced X/Y target judgement
labels. Directions are fitted only on Math or Movies. Blue curves show
$W_{\mathrm{meta}}\!\rightarrow$SJ and brown-orange curves show
$W_{\mathrm{truth}}\!\rightarrow$OC; shading denotes 95\% target-question
cluster-bootstrap intervals. The broad intervals for Llama-3.1-8B and OLMo-3-7B
on MMLU reflect conflict sets with only 13 and 4 B rows, respectively.}
\label{fig:ood-cb-components}
\end{figure}

\section{Implementation and Reproducibility}

\subsection{Models and activation artifacts}

The analyses use four publicly available instruction-tuned language models:
\begin{itemize}
\item Qwen2.5-7B-Instruct
(\texttt{Qwen/Qwen2.5-7B-Instruct});
\item Llama-3.1-8B-Instruct
(\texttt{meta-llama/Llama-3.1-8B-Instruct});
\item Qwen2.5-14B-Instruct
(\texttt{Qwen/Qwen2.5-14B-Instruct}); and
\item OLMo-3-7B-Instruct
(\texttt{allenai/Olmo-3-7B-Instruct}).
\end{itemize}
Activation extraction was performed on a single NVIDIA A800 80GB using bfloat16 forward computation
and float16 activation storage, with a maximum sequence length of 4,096 tokens.
The batch size was four for the 7--8B models in the main-domain extraction and two for Qwen2.5-14B.
OOD activation extraction used a batch size of two.

\begin{table}[!thbp]
\caption{Main-domain activation arrays. Shapes are samples by transformer
blocks by hidden width and apply to each sample-aligned artifact.}
\label{tab:activation-shapes}
\centering
\small
\begin{tabular}{lrrcc}
\toprule
Model & Blocks & Width & Math shape & Movies shape \\
\midrule
Qwen2.5-7B & 28 & 3,584 & $4614\times28\times3584$ & $1824\times28\times3584$ \\
Llama-3.1-8B & 32 & 4,096 & $3078\times32\times4096$ & $5560\times32\times4096$ \\
Qwen2.5-14B & 48 & 5,120 & $4740\times48\times5120$ & $3320\times48\times5120$ \\
OLMo-3-7B & 32 & 4,096 & $724\times32\times4096$ & $1016\times32\times4096$ \\
\bottomrule
\end{tabular}
\end{table}

Each saved artifact contains metadata, a JSONL sample index, and one or more NPZ activation arrays.
The sample index records the activation-row index, question identifier, response, reference answer,
OC label, SJ label, factorial cell, $p_{\mathrm{judge}}$, rendered question-plus-response token count, and prompt hash. Probe fitting, bootstrap inference, and figure generation are performed offline from these saved arrays
and do not require additional model inference.

\section{Interpretive Scope}

The factorial contrasts identify response-level associations with the observed OC and SJ cells;
neither OC nor SJ is experimentally manipulated. Confidence intervals for the primary fixed-window
cross-domain analysis jointly incorporate variation in source- and target-domain question samples,
conditional on the realized pass-at-eight generations and the sampled correct--incorrect response pair
for each eligible question. They do not integrate generation-level or within-question pair-selection variability.
In contrast, the all-layer figures and several secondary controls condition on the fitted
source direction and therefore quantify only target-sample uncertainty.

The OOD tasks use inserted forced-choice responses rather than free generation.
They consequently test whether the fitted directions transfer to a controlled answer site,
not whether the complete free-response generation pipeline replicates across task formats.
In addition, SJ operationalizes correctness-directed self-evaluation under the specified elicitation procedure.
It may contain variance associated with confidence, familiarity, response policy,
or other processes beyond metacognitive accuracy.

Subject to these constraints, the principal pattern is observed across conflict-set ordering,
grouped in-domain validation, cross-domain component AUCs, source-side within-question direction estimation, the all-cell target-question-fixed-effect analysis,
token-count matching, null directions, alternative confidence thresholds, answer-likelihood controls,
and two target tasks excluded from direction fitting.

\end{document}